\documentclass[10pt,a4paper]{article}
\usepackage[scale=.8]{geometry}
\usepackage[affil-it]{authblk}
\usepackage{amssymb,amsmath,amsthm,amsbsy}
\usepackage{physics} 
\usepackage[pdftex,bookmarksnumbered]{hyperref}
\usepackage{graphicx}
\usepackage{epstopdf}
\usepackage{epsfig}
\usepackage{multicol}
\usepackage{subfigure}
\usepackage{multirow}
\usepackage{algorithm}
\usepackage{algpseudocode}
\usepackage{setspace}
\usepackage{color}
\usepackage[dvipsnames]{xcolor}
\usepackage{lineno}
\usepackage{diagbox}
\usepackage{rotating}

\usepackage{tcolorbox}
\usepackage[width=\textwidth]{caption}
\usepackage{lscape}
\usepackage{booktabs}
\usepackage{adjustbox}
\newcommand{\trp}{^{\mathsf{T}}}

\title{\Large\textbf{Gaussian Process Learning of Nonlinear Dynamics}}
\author{Dongwei Ye$^1$ and Mengwu Guo$^{1,2,}$\thanks{Corresponding author (e-mail: \texttt{mengwu.guo@math.lu.se}). The two authors contributed equally to this work.}}
\affil{$^1$Department of Applied Mathematics, University of Twente, the Netherlands \\ \smallskip
$^2$Centre for Mathematical Sciences, Lund University, Sweden}
\date{}

\begin{document}
%\graphicspath{{figures/}}
%\linenumbers

\maketitle % Insert title
\thispagestyle{empty}

\noindent\textit{Abstract}: One of the pivotal tasks in scientific machine learning is to represent underlying dynamical systems from time series data. Many methods for such dynamics learning explicitly require the derivatives of state data, which are not directly available and can be approximated conventionally by finite differences. However, the discrete approximations of time derivatives may result in poor estimations when state data are scarce and/or corrupted by noise, thus compromising the predictiveness of the learned dynamical models. To overcome this technical hurdle, we propose a new method that learns nonlinear dynamics through a Bayesian inference of characterizing model parameters. This method leverages a Gaussian process representation of states, and constructs a likelihood function using the correlation between state data and their derivatives, yet prevents explicit evaluations of time derivatives. Through a Bayesian scheme, a probabilistic estimate of the model parameters is given by the posterior distribution, and thus a quantification is facilitated for uncertainties from noisy state data and the learning process. Specifically, we will discuss the applicability of the proposed method to several typical scenarios for dynamical systems: identification and estimation with an affine parametrization, nonlinear parametric approximation without prior knowledge, and general parameter estimation for a given dynamical system.

\vspace{1mm}

\section{Introduction}

Data-driven learning of dynamical systems from time series is an important component in scientific machine learning, as it bridges the gap between data-driven approximation and physics-based modeling. Such a learning task is often posed as an inverse problem with given parametrization of the dynamical system \cite{ghattas2021learning,willcox2021imperative}. Without loss of generality, therefore, we consider the following representation of nonlinear dynamical systems:
\begin{equation}\label{eq:system}
\begin{cases}
\dot{x}_1(t) = f_1(\vb*{x}(t);\vb*{\theta}_1)\\
\dot{x}_2(t) = f_2(\vb*{x}(t);\vb*{\theta}_2) \\
\cdots~\cdots\\
\dot{x}_N(t) = f_N(\vb*{x}(t);\vb*{\theta}_N)
\end{cases}
\text{with} \quad \vb*{x}(t_0) = \vb*{x}_0\,, \quad t \geq t_0\,,
\end{equation}
in which $\vb*{x}=\{x_1,x_2,\cdots,x_N\}\trp\in \mathbb{R}^N$ collects the $N$ state variables in the system, $\vb*{x}_0$ specifies the initial condition at $t=t_0$, and $\{f_i(\cdot;\vb*{\theta}_i)\}_{i=1}^N$ define the system parametrization and are not-all-linear functions, the $i$-th of which is characterized by a collection of $p_i$ unknown parameters denoted by $\vb*{\theta}_i\in\mathbb{R}^{p_i}$. Hence, parameter estimation/inference methods can be used to determine the values of $\{\vb*{\theta}_i\}_{i=1}^{N}$ with a set of solution data $\{\vb*{x}(t_k) \}_{k=0}^{K-1}$ collected at $K$ time-instances $\mathcal{T} = \{t_k\}_{k=0}^{K-1}$ ($t_0 < t_1 < \cdots <t_{K-1}$). 

Ideally, physical consistency should be preserved in the learning process through a proper system parametrization, so that non-physical behaviors in the data are filtered out and reliability in long-term predictions is achieved. One popular choice is the sparse identification of nonlinear dynamics (SINDy) \cite{Steven2016}, which defines $f_i(\cdot;\vb*{\theta}_i)$ as a linear combination of candidate functions and uses sparse regression to determine active terms among them. The effectiveness of identification in this method relies on the inclusiveness of candidates and the assumption of sparsity. Relevantly, data-driven operator inference \cite{Benjamin2016,Qian2022,guo2022bayesian} focuses on reduced-order dynamics learning, and attempts to recover a low-dimensional system \eqref{eq:system} with a polynomial structure $f_i(\cdot;\vb*{\theta}_i)$ inspired by known governing equations. Formulated as (Tikhonov-regularized) least squares \cite{mcquarrie2021data}, such a non-intrusive method seeks to find estimators of $\vb*{\theta}_i$ that best match the training data in a minimum-residual sense, and hence avoids entailing access to source code. This is especially advantageous for applications with readily-executed solvers that cannot be easily modified.
Alternative to pinpointing the exact structure of a dynamical system, $f_i(\cdot;\vb*{\theta}_i)$ can be considered as an approximation of the right-hand side of the differential equation with parameters $\vb*{\theta}_i$. For example, neural ordinary differential equations \cite{Chen2018} adopt a multi-layer perceptron network as $f_i(\cdot;\vb*{\theta}_i)$, while the Runge-Kutta neural network method \cite{zhuang2021model} constructs parametric surrogate models for the discrete-time integrator. 

For many of these dynamics learning techniques, however, predictive performance can be compromised when training data are scarce and/or corrupted by noise. In particular with the `equation-fitting' strategy, i.e., minimum-residual of the continuous- or discrete-time formulation of \eqref{eq:system}, there typically involve time derivative approximations by numerical differentiation (e.g., finite differences). The data noise and/or sparsity may lead to a poor estimation of time derivatives and then curtail the accuracy of parameter inference. Efforts have been made to avoid derivative evaluations by considering the misfit in state data (`trajectory-fitting') directly instead of minimizing residuals, such as incorporating time integration into loss functions \cite{Chen2018} or as constraints \cite{uy2023operator}. However, these settings typically demand more sophisticated numerical treatments, which are not always convenient, flexible, or effective for complex dynamical behaviors. Meanwhile, particularly for the cases with imperfect datasets, it is crucial to facilitate the learning process with uncertainty quantification and enable a model validation in terms of accuracy and robustness.
A straightforward strategy is ensemble learning that leverages bootstrap sampling of time-series data to aggregate statistical estimates from multiple models \cite{Fasel2022}, but the many times of evaluations may escalate computational costs. An efficient and rigorous alternative is Bayesian inference \cite{box2011bayesian} for system parameters $\vb*{\theta}_i$, deemed capable of taking data and modeling uncertainties into systematical consideration \cite{Naoya2017,zhang2018robust,yang2021inference,Hirsh2022,guo2022bayesian}. In this work, we aim to address the aforementioned two challenges -- robustness with scarce/noisy data and capability of uncertainty quantification -- through a novel Bayesian inference method for learning dynamical systems \eqref{eq:system}. The primary novelty lies in integrating Gaussian process modeling into Bayesian inference to smooth and/or interpolate imperfect data, accounting for the approximation and uncertainty estimation of time derivatives, and incorporating differential equation constraints into the inference process.

Gaussian process emulation is a probabilistic approach to supervised learning \cite{williams2006gaussian}, and its variations have been applied to a wide range of computational tasks in science and engineering, see \cite{Bonilla2007,MARREL2009742,Damianou2013,Liu2014,Chang2015,costabal2019multi,Ye2022,Botteghi2022,cicci2023} for a few examples. 
More relevant to this work is the fact that Gaussian processes are also capable of embedding differential equation constraints into kernel structures. Specifically, as Gaussianity can be preserved through linear operations, the latent force models \cite{Alvarez2009} for Gaussian processes have paved the way for describing the correlations between solution states/fields and their latent `forces' through linear differential operators. 
This idea has been used for solving forward and inverse problems governed by linear partial differential equations \cite{RAISSI2017683,pfortner2022physics}, and extended to nonlinear equations \cite{Yifan2021,meng2023sparse} with error analysis \cite{batlle2023error}, as well as to nonlinear operator learning \cite{batlle2023kernel}. 
Relevant techniques have been explored in various contexts in scientific computing, such as stochastic partial differential equations \cite{Simo2011}, statistical finite element analysis \cite{Mark2021}, port-Hamiltonian systems \cite{beckers2022gaussian}, and conservation laws \cite{hansen2023learning}.

In this work, a vector-valued Gaussian process is constructed to represent the statistical correlation between the states and their time derivatives. Combined with the system parametrization $f_i(\cdot;\vb*{\theta}_i)$, this correlation is leveraged to formulate a novel likelihood function constrained by the differential equations \eqref{eq:system}, in addition to a prior distribution of the parameters $\vb*{\theta}_i$ needed for the Bayesian inference. Thereafter, a probabilistic estimate of $\vb*{\theta}_i$ is given by the posterior distribution through the Bayes' rule. By sampling over this distribution, uncertainties are consequently propagated into the predictions produced by the learned dynamical system, enabling a predictive uncertainty quantification. Importantly, the Gaussian process representation also provides a natural smoother by kernel regression, which counters the unsatisfactory accuracy in time derivatives caused by noisy/scarce state data. We will further explain the proposed method in two scenarios: one with a given affine structure of the dynamical system, and the other for a general nonlinear  approximation of the system without prior knowledge. 

Following the introduction, the proposed Gaussian-process-based Bayesian inference method for learning nonlinear dynamics is presented in Section 2. The two aforementioned scenarios are discussed in Section 3, for which numerical results are given in Section 4. Concluding remarks are finally made in Section 5.

\section{Dynamics learning with Gaussian processes}

Introduced in this section are the core ideas on using Gaussian processes to estimate the system parameters in \eqref{eq:system} in a Bayesian framework. Assuming that the parameter estimation in each equation can be conducted independently, we generically consider the Bayesian inference for $\vb*{\theta}_i$ in the $i$-th equation, $1\leq i \leq N$. In this work, uncorrelated Gaussian distribution is adopted as the prior for the model parameters, written as
\begin{equation}
\pi_\texttt{prior}(\vb*{\theta}_i) \propto \exp\left(-\frac{1}{2}\|\vb*{\theta}_i\|_{\vb*{\Lambda}_i}^2\right)\,,
\end{equation}
in which $\|\vb*{\theta}_i\|_{\vb*{\Lambda}_i}:= \sqrt{\vb*{\theta}_i\trp \vb*{\Lambda}_i \vb*{\theta}_i}$ with $\vb*{\Lambda}_i := \text{diag}(\lambda_{i1},\cdots, \lambda_{ip_i})$, and $\lambda_{ij}\geq 0$ over $1\leq j \leq p_i$. To apply the Bayes' rule, a likelihood needs to be defined for the training data.

\subsection{Likelihood definition with a Gaussian process}

Here we assume that each component $x_i(t)$ ($1\leq i \leq N$) of the state vector $\vb*{x}(t)$ follows a zero-mean Gaussian process over the time $t$:
\begin{equation}
x_i(t) \sim \mathcal{GP}(0,\kappa_i(t,t'))\,,
\end{equation}
in which $\kappa_i$ is a given smooth kernel function representing the covariance over the Gaussian process, i.e., $\kappa_i(t,t') = \mathbb{C}ov[x_i(t),x_i(t')]$. Particularly, we use the squared exponential kernel as the covariance function in this work. Thus, the time-derivative of $x_i(t)$ also follows a Gaussian process, and the correlation between the state $x_i$ and its time-derivative $\dot{x}_i$ is given by
\begin{equation}\label{eq:jointfunc}
\begin{bmatrix}
x_i(t) \\ \dot{x}_i(t)
\end{bmatrix} \sim \text{vec-}\mathcal{GP}\left(
\begin{bmatrix}
0 \\ 0
\end{bmatrix}, 
\begin{bmatrix}
 \kappa_i(t,t')  &  \partial_{t'}\kappa_i(t,t') \\
\partial_{t}\kappa_i(t,t') & \partial_{t}\partial_{t'}\kappa_i(t,t') 
\end{bmatrix}
\right)\,.
\end{equation}
Here $\text{vec-}\mathcal{GP}$ stands for a vector-valued Gaussian process.

% (i.e., $\{x_i(t_k)\}_{k=0}^{K-1}$)

For notation, we collect the observables of $x_i(t)$ at time-instances $\mathcal{T}$ in a size-$K$ random vector $\vb*{U}_i$, while their observed values correspondingly in $\vb{u}_i\in \mathbb{R}^K$, let $\vb*{D}_i$ denote the size-$K$ random vector of $x_i$'s time derivatives over $\mathcal{T}$, and let $\vb{Z} = [\vb{u}_1 ~\cdots ~ \vb{u}_N]\trp = [\vb*{x}(t_0)~\cdots ~ \vb*{x}(t_{K-1})]\in \mathbb{R}^{N\times K}$ collect the full-state data of $\vb*{x}(t)$ over $\mathcal{T}$. As a Gaussian process assumption on $x_i(t)$ has been adopted, $\vb*{D}_i$ combined with $\vb*{U}_i$ should follow a joint normal distribution written as
\begin{equation}\label{eq:jointdis}
\begin{split}
\begin{bmatrix}
\boldsymbol{D}_i \\ \boldsymbol{U}_i
\end{bmatrix} & \sim \mathcal{N}\left(
\begin{bmatrix}
\mathbf{0} \\ \mathbf{0}
\end{bmatrix},
\begin{bmatrix}
\mathbf{K}^{dd}_i & \mathbf{K}^{du}_i \\
\mathbf{K}^{ud}_i & \mathbf{K}^{uu}_i
\end{bmatrix}
\right)\,,\quad \text{with} \\
\vb{K}_i^{dd} & =  \partial_{t}\partial_{t'}\kappa_i(\mathcal{T},\mathcal{T}) + \chi^d_i\vb{I}_{K}\in \mathbb{R}^{K\times K} \,,\\
\vb{K}_i^{uu} & =  \kappa_i(\mathcal{T},\mathcal{T}) + \chi^u_i\vb{I}_{K}\in \mathbb{R}^{K\times K} \,,\\
\vb{K}_i^{du} & =  \partial_{t}\kappa_i(\mathcal{T},\mathcal{T}) = (\vb{K}_i^{ud})\trp \in \mathbb{R}^{K\times K}\,.
\end{split}
\end{equation}
Considering the possibility that the data $\vb{Z}$ may be corrupted by noise, we have added white noise terms with variance values $\chi^u_i>0$ and $\chi^d_i>0$ to $\vb*{U}_i$ and $\vb*{D}_i$, respectively. 
Furthermore, we define an additional random vector $\vb*{\delta}_i = \vb*{D}_i  - f_i(\vb{Z};\vb*{\theta}_i)$ that represents the residuals over $\mathcal{T}$ with given state data $\vb{Z}$ and a fixed $\vb*{\theta}_i$.

In this work, the core idea is to construct a likelihood function not only on the direct observations, but also on the alignment between $\vb*{D}_i$ and $f_i(\vb{Z};\vb*{\theta}_i)$, i.e., $\vb*{\delta}_i = \vb{0}$, the satisfaction of the dynamical system formulation \eqref{eq:system}. Such a likelihood function can thus be formulated from \eqref{eq:jointdis} as follows:
\begin{equation}\label{eq:likelihood}
\begin{split}
\pi_\texttt{like}(\vb*{\theta}_i) & = p\left(\vb*{\delta}_i=\vb{0},\vb*{U}_i = \vb{u}_i\mid \vb{Z},\vb*{\theta}_i\right) \\
& \propto  \exp\left(-\frac{1}{2}
\begin{bmatrix}
f_i(\vb{Z};\vb*{\theta}_i) \\ \vb{u}_i
\end{bmatrix}\trp
\begin{bmatrix}
\vb{K}^{dd}_i & \vb{K}^{du}_i \\
\vb{K}^{ud}_i & \vb{K}^{uu}_i
\end{bmatrix}^\text{-1}
\begin{bmatrix}
f_i(\vb{Z};\vb*{\theta}_i) \\ \vb{u}_i
\end{bmatrix}
\right)\,.
\end{split}
\end{equation}
 Therefore, in addition to the data-driven nature, such a likelihood definition guarantees a physics-based interpretation.

\medskip\noindent {\bfseries Remark 1}: In this work, we consider both $\vb*{U}_i$ and $\vb*{D}_i$ over the whole data coverage $\mathcal{T}$, but this is not a requirement, i.e., $\vb*{U}_i$ and $\vb*{D}_i$ can correspond to two different subsets of time instances in $\mathcal{T}$.

\subsection{Posterior distribution for Bayesian inference}

Once the prior and likelihood are defined, the Bayes’ rule gives the posterior distribution
\begin{equation}\label{eq:inference}
\begin{split}
\pi_\texttt{post}(\vb*{\theta}_i) = p\left(\vb*{\theta}_i \mid \vb*{\delta}_i = \vb{0}, ~\vb*{U}_i = \vb{u}_i, ~\vb{Z}\right)   ~& = \frac{\pi_\texttt{like}(\vb*{\theta}_i) \pi_\texttt{prior}(\vb*{\theta}_i)}{\int \pi_\texttt{like}(\vb*{\theta}_i) \pi_\texttt{prior}(\vb*{\theta}_i) \dd\vb*{\theta}_i} \\
~& \propto \pi_\texttt{like}(\vb*{\theta}_i) \pi_\texttt{prior}(\vb*{\theta}_i)\,.
\end{split}
\end{equation}
Note that the inverse of the kernel matrix in the likelihood \eqref{eq:likelihood} can be rewritten as
\begin{equation}
\begin{split}
\begin{bmatrix}
\vb{K}^{dd}_i & \vb{K}^{du}_i \\
\vb{K}^{ud}_i & \vb{K}^{uu}_i
\end{bmatrix}^\text{-1} & = 
\begin{bmatrix}
\vb{R}^{dd}_i & \vb{R}^{du}_i \\
\vb{R}^{ud}_i & \vb{R}^{uu}_i
\end{bmatrix} \,, \quad \text{with}\\
\vb{R}^{dd}_i & = \left(\vb{K}^{dd}_i - \vb{K}^{du}_i(\vb{K}^{uu}_i)^{-1}\vb{K}^{ud}_i \right)^{-1} \,,\\
\vb{R}^{uu}_i & = \left(\vb{K}^{uu}_i - \vb{K}^{ud}_i(\vb{K}^{dd}_i)^{-1}\vb{K}^{du}_i \right)^{-1} \,,\\
\vb{R}^{du}_i & = -\vb{R}^{dd}_i\vb{K}^{du}_i (\vb{K}^{uu}_i)^{-1}= (\vb{R}^{ud}_i)\trp\,.
\end{split}
\end{equation}
By substituting this into the $\pi_\texttt{like}$ expression \eqref{eq:likelihood} that is used in the inference \eqref{eq:inference}, the posterior distribution is therefore written explicitly as 
\begin{equation}\label{eq:posterior}
\begin{split}
& ~~\pi_\texttt{post}(\vb*{\theta}_i) \propto \pi_\texttt{like}(\vb*{\theta}_i) \pi_\texttt{prior}(\vb*{\theta}_i)\\
 \propto  & ~
\exp\left(-\frac{1}{2}\left(f_i(\vb{Z};\vb*{\theta}_i)\trp\vb{R}_i^{dd}f_i(\vb{Z};\vb*{\theta}_i) +2f_i(\vb{Z};\vb*{\theta}_i)\trp\vb{R}_i^{du}\vb{u}_i + \|\vb*{\theta}_i\|_{\vb*{\Lambda}_i}^2 \right)\right) \,.
\end{split}
\end{equation}
Additionally, we define an estimate of the time derivatives over $\mathcal{T}$ (i.e., $\dot{x}_i(\mathcal{T})$) using a Gaussian process regression with the state observations $(\mathcal{T},\vb{u}_i)$ at the same instances, given as
\begin{equation}\label{eq:derivative}
    \hat{\vb{d}}_i:= \vb{K}^{du}_i (\vb{K}^{uu}_i)^{-1}\vb{u}_i\,.
\end{equation}
These are the time derivatives extracted from the mean function $\kappa_i(t,\mathcal{T})(\vb{K}_i^{uu})^{-1}\vb{u}_i$ of the Gaussian process reconstruction.
Considering the fact that $\vb{R}_i^{dd}\hat{\vb{d}}_i = -\vb{R}_i^{du}\vb{u}_i$, the posterior expression in \eqref{eq:posterior} is hence rearranged into
\begin{equation}\label{eq:posterior1}
    \pi_\texttt{post}(\vb*{\theta}_i) \propto \exp\left(-\frac{1}{2}\left(\left\|f_i(\vb{Z};\vb*{\theta}_i)-\hat{\vb{d}}_i\right\|_{\vb{R}_i^{dd}}^2 + \|\vb*{\theta}_i\|_{\vb*{\Lambda}_i}^2 \right)\right)\,, 
\end{equation}
where $\left\|\bullet\right\|_{\vb{R}_{i}^{dd}} := \sqrt{\bullet\trp\vb{R}_{i}^{dd}\bullet}$. It is worth noting that, in both \eqref{eq:posterior} and \eqref{eq:posterior1}, there are  constant terms (i.e., independent of $\vb*{\theta}_i$) omitted in the exponential functions.

One can choose suitable numerical techniques, e.g., Markov chain Monte Carlo or variational inference, to approximate this posterior distribution with specific parametrization of $f_i(\cdot;\vb*{\theta}_i)$. We also note that an estimator of \emph{maximum a posteriori} (MAP) from \eqref{eq:posterior1} is
\begin{equation}\label{eq:MAP_LS}
    (\vb*{\theta}_i)_\text{MAP} = \arg\min_{\vb*{\theta}_i\in \mathbb{R}^{p_i}} \left\{\left\|f_i(\vb{Z};\vb*{\theta}_i)-\hat{\vb{d}}_i\right\|_{\vb{R}_i^{dd}}^2 + \|\vb*{\theta}_i\|_{\vb*{\Lambda}_i}^2\right\}\,.
\end{equation}
This is a regularized, generalized least squares estimator of $\vb*{\theta}_i$ weighted by $\vb{R}_i^{dd}=(\vb{K}^{dd}_i - \vb{K}^{du}_i(\vb{K}^{uu}_i)^{-1}\vb{K}^{ud}_i)^{-1}$, the inverse of which is the posterior covariance matrix of $\dot{x}_i(\mathcal{T})$ through Gaussian process regression. 
Inside the first term of the loss function is the residual of the governing differential equation over $\mathcal{T}$ with time derivatives approximated by Gaussian process regression. Essentially, this confirms the interpretability of the proposed method. With the weight matrix $\vb{R}_i^{dd}$, the uncertainty introduced by the derivative approximation is included into the posterior estimation.

\medskip\noindent {\bfseries Remark 2}: A physically meaningful parametrization for the dynamical system \eqref{eq:system} is independent of initial conditions, so should the inference of parameters $\vb*{\theta}_i$ be. The inference can be performed with time series data collected for multiple initial conditions, in which case \eqref{eq:posterior1} becomes
\begin{equation}
    \pi_\texttt{post}(\vb*{\theta}_i) 
    \propto  \exp\left(-\frac{1}{2}\left(\sum_\texttt{all ICs}\left\{\left\|f_i(\vb{Z};\vb*{\theta}_i)-\hat{\vb{d}}_i\right\|_{\vb{R}_i^{dd}}^2 \right\}_\texttt{each IC} + \|\vb*{\theta}_i\|_{\vb*{\Lambda}_i}^2\right)\right)\,,
\end{equation}
where the loss terms $\|f_i(\vb{Z};\vb*{\theta}_i)-\hat{\vb{d}}_i\|_{\vb{R}_i^{dd}}^2$ are summed over all the initial conditions (ICs) included in the dataset. Note in this case that the trajectory with each initial condition is approximated by an individual GP, leading correspondingly to an individual $\vb{R}_i^{dd}$ as well. In particular, when data with a single initial condition are scarce, one can try to improve the parameter inference by adding data with other initial conditions. The estimated parameter values should work for the predictions for new initial conditions outside the coverage of training data.

\medskip\noindent {\bfseries Remark 3}: In the beginning of this section, we assumed an independent parameter estimation for each equation in \eqref{eq:system}. Therefore, if multiple equations share a parameter, there are multiple estimates of this parameter individually derived from the involved equations. It is most possible that these estimates do not coincide with each other perfectly. In this case, we can jointly infer the parameters from all these equations to ensure a single, informative estimate of the shared parameter(s). For example, if $\vb*{\theta}_m$ and $\vb*{\theta}_n$ ($m\neq n$) have shared parameters, i.e., $\vb*{\theta}_m\cap \vb*{\theta}_n \neq \emptyset$, we can jointly infer $\vb*{\theta}_m\cup \vb*{\theta}_n$ as
\begin{equation}\label{eq:shareparam} 
\begin{split}
    & \pi_\texttt{post}(\vb*{\theta}_m\cup \vb*{\theta}_n) = p\left(\vb*{\theta}_m\cup \vb*{\theta}_n \mid \vb*{\delta}_m = \vb{0}, ~\vb*{\delta}_n = \vb{0}, ~\vb*{U}_m = \vb{u}_m, ~\vb*{U}_n = \vb{u}_n, ~\vb{Z}\right) \\
    \propto & \exp\left(-\frac{1}{2}\left(\left\|f_m(\vb{Z};\vb*{\theta}_m)-\hat{\vb{d}}_m\right\|_{\vb{R}_m^{dd}}^2 + \left\|f_n(\vb{Z};\vb*{\theta}_n)-\hat{\vb{d}}_n\right\|_{\vb{R}_n^{dd}}^2 + \|\vb*{\theta}_i\|_{\vb*{\Lambda}_m}^2 + \|\vb*{\theta}_i\|_{\vb*{\Lambda}_n}^2 \right)\right)\,.
\end{split} 
\end{equation}

\subsection{Bayesian predictions}

As we have treated the Bayesian inference of each $\vb*{\theta}_i$ independently, $1\leq i \leq N$, the joint posterior distribution of all unknown parameters is
\begin{equation} 
\pi_\texttt{post}(\vb*{\theta}_1, \cdots, \vb*{\theta}_N ) = p\left(\vb*{\theta}_1, \cdots, \vb*{\theta}_N\mid \left\{\vb*{\delta}_i = \vb{0}, ~\vb*{U}_i = \vb{u}_i\right\}_{i=1}^{N},\vb{Z} \right) = \prod_{i=1}^{N} \pi_\texttt{post}(\vb*{\theta}_i)\,. 
\end{equation}
Given $(\vb*{\theta}_1, \cdots, \vb*{\theta}_N)$, we can solve for the states $\vb*{x}(t)$ through the dynamical system \eqref{eq:system} for $t>t_0$. The states can therefore be viewed as a stochastic processes because they depend on the random variables $(\vb*{\theta}_1, \cdots, \vb*{\theta}_N)$
with distribution
\begin{equation}\label{eq:Bayes_prediction} 
p\left(\vb*{x}(t)\mid \left\{\vb*{\delta}_i = \vb{0}, ~\vb*{U}_i = \vb{u}_i\right\}_{i=1}^{N},\vb{Z}\right) = \int p(\vb*{x}(t) | \vb*{\theta}_1, \cdots, \vb*{\theta}_N ) \prod_{i=1}^{N} \pi_\texttt{post}(\vb*{\theta}_i)\,, 
\end{equation}
in which all $\vb*{\theta}_i$'s are marginalized. For instance, Monte Carlo sampling can be employed over the posterior $\pi_\texttt{post}(\vb*{\theta}_1, \cdots, \vb*{\theta}_N )$ to estimate the mean function and second-order moments of $\vb*{x}(t)$. In fact, $p(\vb*{x}(t) | \vb*{\theta}_1, \cdots, \vb*{\theta}_N )$ represents a deterministic dynamical model, and one
can hence write the state solution $\vb*{x}(t)$ determined by the parameters $(\vb*{\theta}_1, \cdots, \vb*{\theta}_N)$ as a function of them, i.e., $
\vb*{x}(t; \vb*{\theta}_1, \cdots, \vb*{\theta}_N)$.

\subsection{Implementation considerations}

\noindent \textbf{Construction of} $f_i({\vb{U}};\vb*{\theta}_i)$

\medskip\noindent To further reduce the impact of data noise, we can employ a corrected version of $f_i(\hat{\vb{Z}};\vb*{\theta}_i)$ with smoothed state data $\hat{\vb{Z}}= [\hat{\vb{u}}_i ~\cdots ~ \hat{\vb{u}}_N]\trp$ through Gaussian process reconstruction, i.e., 
\begin{equation}
    \hat{\vb{u}}_i = \kappa_i(\mathcal{T},\mathcal{T})(\vb{K}_i^{uu})^{-1}\vb{u}_i\,.
\end{equation}
This treatment can smooth the noise-corrupted trajectories through kernel regression and practically improves the approximation of $f_i$, which further guarantees the quality of Bayesian inference.

\medskip\noindent \textbf{Hyperparameter tuning}

\medskip\noindent There are hyperparameters throughout the Gaussian process modeling, including kernel parameters in $\kappa_i$, the noise variance $\chi_i^u$ and $\chi_i^{d}$ respectively for the states and derivatives, and the regularization coefficients $\vb*{\lambda}_i$. In principle, their values can be empirically determined by maximizing the marginal likelihood \cite{williams2006gaussian}, i.e.,  $\int \pi_\texttt{like}(\vb*{\theta}_i) \pi_\texttt{prior}(\vb*{\theta}_i) \dd\vb*{\theta}_i$. However, this marginal likelihood function is often intractable to compute in practice, which makes the optimization challenging or even impossible, and there is little guarantee that the learned dynamical models are accurate and stable with these hyperparameter values. Instead, we suggest the following pragmatic strategy:
\begin{enumerate}
    \item Kernel parameters and noise term $\chi_i^u$: we simply consider the marginal likelihood function $p(\vb{u}_i)$ for $\vb*{U}_i$ only (with the $\vb*{D}_i$ term marginalized). Given in the closed form
    \begin{equation}
    \log p(\vb{u}_i) = -\frac{1}{2}\vb{u}_i\trp(\vb{K}_i^{uu})^{-1}\vb{u}_i - \frac{1}{2}\log |\vb{K}_i^{uu}| - \frac{K}{2}\log 2\pi\,,
    \end{equation}
    this marginal likelihood function only involves the kernel parameters and $\chi_i^u$ and can be maximized to determine their values.
    It is worth pointing out that the Gaussian process hyperparameters are determined individually to adapt to the amplitude and lengthscale of each $x_i$, and thus the covariance is not constant across the components of the state vector $\vb*{x}$.
    \item Noise term $\chi_i^d$: the variance value $\chi_i^{d}$ for the additive noise onto the time derivatives is adjusted to ensure that $(\vb{R}_i^{dd})^{-1}= \partial_{t}\partial_{t'}\kappa_i(\mathcal{T},\mathcal{T}) + \chi^d_i\vb{I}_{K} - \vb{K}^{du}_i(\vb{K}^{uu}_i)^{-1}\vb{K}^{ud}_i$ is well-conditioned, which is a numerical requirement for evaluating $\vb{R}_i^{dd}$ needed in the Bayesian inference. On the other hand, as $\chi_i^d$ measures the discrepancy between $\vb{d}_i$ and $f_i(\vb{Z};\vb*{\theta}_i)$, it is a valid assumption that $\chi_i^d$ should not be a large value when the smoothed $f_i(\hat{\vb{Z}};\vb*{\theta}_i)$ is used, for which we suggest $\chi_i^d$ and $\chi_i^u$ share the same order of magnitude.
    \item Coefficients $\vb*{\Lambda}_i$ in regularization: we will discuss the settings of $\lambda_{ij}$'s later in typical scenarios of dynamics learning in Section 3, and more specifically in the numerical experiments in Section 4. 
\end{enumerate}

\section{Two scenarios of system parametrization}
In this section, to demonstrate the generality of the proposed Bayesian inference method for dynamics learning, we discuss two scenarios for the $\vb*{\theta}_i$-parametrization of $f_i(\cdot;\vb*{\theta}_i)$ in \eqref{eq:system}. The first scenario considers cases where the dynamical system's structure is given in an affine form, and the estimation of system parameters becomes a regularized linear regression problem. On the contrary, the second scenario assumes no prior knowledge on the system structure, and instead uses a general, nonlinear parametrization to approximate. 

\subsection{Scenario (I): linear parametrization}

In the first scenario, we consider an affine parametrization of $f_i(\cdot;\vb*{\theta}_i)$ that is linear with respect to $\vb*{\theta}_i$, i.e., 
\begin{equation}\label{eq:linear}
f_i(\vb*{x};\vb*{\theta}_i) = \vb*{g}_i(\vb*{x})\trp\vb*{\theta}_i\,,
\end{equation}
in which $\vb*{g}_i: \mathbb{R}^N \to \mathbb{R}^{p_i}$ is a given vector-valued function. $f_i(\vb{Z};\vb*{\theta}_i)$ is hence rewritten as $\vb{G}_i\vb*{\theta}_i$ with $\vb{G}_i:= [\vb*{g}_i(\vb*{\hat{x}}(t_0))~\cdots ~ \vb*{g}_i(\vb*{\hat{x}}(t_{K-1}))]\trp \in\mathbb{R}^{K\times p_i}$, a linear transformation acting on the parameter vector $\vb*{\theta}_i$. Thus, the posterior \eqref{eq:posterior1} given by the proposed inference method is explicitly represented as
\begin{equation}\label{eq:L2posterior}
\begin{split}
\pi_\texttt{post}(\vb*{\theta}_i) & \propto 
\exp\left(-\frac{1}{2}\left(\left\|\vb{G}_i\vb*{\theta}_i-\hat{\vb{d}}_i\right\|_{\vb{R}_i^{dd}}^2 +\|\vb*{\theta}_i\|_{\vb*{\Lambda}_i}^2 \right)\right) \\
& \propto \exp\left(-\frac{1}{2}(\vb*{\theta}_i-\vb*{\mu}_i)\trp \vb*{\Sigma}_i^{-1}(\vb*{\theta}_i-\vb*{\mu}_i)\right)\,, 
\end{split}
\end{equation}
i.e., the posterior of $\vb*{\theta}_i$ follows a joint normal distribution $\mathcal{N}(\vb*{\mu}_i,\vb*{\Sigma}_i)$, in which the mean vector $\vb*{\mu}_i$ coincides with an MAP estimator written as
\begin{equation}\label{eq:L2estimate}
\vb*{\mu}_i=\left(\vb*{\theta}_i\right)_\text{MAP}  = (\vb{G}_i\trp\vb{R}_i^{dd}\vb{G}_i+\vb*{\Lambda}_i)^{-1} \vb{G}_i\trp\vb{R}^{dd}_i \hat{\vb{d}}_i\,,
\end{equation}
and the posterior covariance matrix is
\begin{equation}\label{eq:sc1_cov}
\vb*{\Sigma}_i =(\vb{G}_i\trp\vb{R}_i^{dd}\vb{G}_i + \vb*{\Lambda}_i)^{-1}\,.
\end{equation}
In fact, the Gaussian prior of $\vb*{\theta}_i$ provides a Tikhonov regularization here, and the estimation of $\vb*{\theta}_i$ becomes a ridge regression \cite{murphy2012machine}.

\medskip\noindent \textbf{Case A: parameter estimation}

\medskip
If the system structure \eqref{eq:linear} is known and all the components in $\vb*{g}_i(\vb*{x})$ are active, the determination of the parameters $\vb*{\theta}_i$ is a simple linear regression problem. In this case, we set $\vb*{\Lambda}_{i}=\lambda_i\vb{I}_{p_i}$, i.e., the parameters follow independent and identically distributed Gaussian priors. When $\lambda_i=0$, i.e., the prior variance is infinitely large and the distribution thus becomes uninformative, the corresponding Tikhonov regularization vanishes. This still works for the proposed inference if $\vb{G}_i$ has full column rank and $\vb{G}_i\trp\vb{R}_i^{dd}\vb{G}_i$ is well-conditioned.

\medskip\noindent \textbf{Case B: sparse identification}

\medskip
Alternatively, we may consider that $\vb*{g}_i(\vb*{x})$ collects a set of candidate terms and only a few of them are active. Though the linear parametrization given in \eqref{eq:linear} remains, there is an additional assumption of sparsity in $\vb*{\theta}_i$, i.e., many components of $\vb*{\theta}_i$ should be zero. In this case, we aim to use the proposed inference technique to identify the active candidates with time-series data \cite{Steven2016}.

In the Bayesian inference of \ref{eq:L2posterior}, each penalty coefficient $\lambda_{ij}$ in $\vb*{\Lambda}_i$ is inversely proportional to the variance of the corresponding Gaussian prior for $\theta_{ij}$ ($j$th entry of $\vb*{\theta}_i$). In other words, the magnitude of $\lambda_{ij}$ reflects the `confidence' in the given prior, and this will further influence the Bayesian estimation, especially the reflection of sparsity promotion in uncertainty quantification. In this paper, we suggest the following steps to set $\lambda_{ij}$ values for each equation $i$:
\begin{enumerate}
    \item Sequential threshold ridge regression \cite{thaler2019sparse} is first applied to the following minimization problem
    \begin{equation}\label{eq:sqtrr}
        \min_{\vb*{\theta}_i\in \mathbb{R}^{p_i}}\left\{\|\vb{G}_i\vb*{\theta}_i-\hat{\vb{d}}_i\|_2^2 + \|\vb*{\theta}_i\|_2^2\right\}\,,
    \end{equation}
    so that the active $\theta_{ij}$ terms over the $j$-indices are found and collected in $\vb*{\theta}_i^\text{a}$, while the other (sparsified) terms in $\vb*{\theta}_i^\text{s}$. Note that an $L^2$-regularized original least squares problem is considered here for an efficient process of sequential threshold truncation. 
    \item With $\theta_{ij}$ terms divided into two groups, we define $\|\vb*{\theta}_i\|_{\vb*{\Lambda}_i}^2= \sum_{j=1}^{p_i}\lambda_{ij}|\theta_{ij}|^2:= \lambda_i^\text{a}\|\vb*{\theta}_i^\text{a}\|_2^2+ \lambda_i^\text{s}\|\vb*{\theta}_i^\text{s}\|_2^2$, i.e., a single $\lambda$-value is shared among the $\theta_{ij}$'s in the same group.
    \item Values are then assigned to $\lambda_i^\text{a}$ and $\lambda_i^\text{s}$ such that $\lambda_i^\text{s}>1>\lambda_i^\text{a}> 0$ and $\lambda_i^\text{s} \gg \lambda_i^\text{a}$. Here a large value of $\lambda_i^\text{s}$ leads to a sharp normal distribution centered at 0, indicating concrete confidence in the sparsity inferred through the sequential threshold ridge regression.
\end{enumerate}

\subsection{Scenario (II): nonlinear approximation with a shallow neural network}

\noindent When there is no prior knowledge available for the dynamical system to be learned, a commonly used strategy is to approximate $f_i(\cdot;\vb*{\theta}_i)$ with a nonlinear regression model parametrized by $\vb*{\theta}_i$. In the second scenario, we hence adopt the spirit of neural ordinary differential equations \cite{Chen2018}, and use a shallow neural network with a single hidden layer to parametrize $f_i(\cdot;\vb*{\theta}_i)$. Specifically, we let 
\begin{equation}
f_i(\vb*{x};\vb*{\theta}_i) = \sum_{l=1}^{L}v_{il} ~\sigma\left( \vb*{w}_{il}\trp \vb*{x} + b_{il} \right)\,,\quad \text{with}~~
\vb*{\theta}_i := \{v_{il},b_{il},\vb*{w}_{il}\}_{l=1}^{L}\,,
\end{equation}
in which $\vb*{w}_{il}\in \mathbb{R}^N$ stands for the weights between the input and hidden layers, $v_{il}$'s are those between the hidden and output layers, $b_{il}$'s are the biases at the hidden neurons, and there are $L$ neurons in the hidden layer. In general, such a neural network parametrization features a strong expressive power for nonlinearity. With a Gaussian prior defined on the network parameters $\vb*{\theta}_i$, we resort to Markov chain Monte Carlo for the Bayesian inference of neural network parameters and use the Metropolis–Hastings algorithm to achieve the posterior \eqref{eq:posterior1}. 

\section{Numerical experiments}
\subsection{Lotka–Volterra equations}
Lotka-Volterra equations describe a nonlinear system often used to model the population dynamics between two species in an ecological system. The equations are given as
\begin{equation}
\begin{cases}
\dot{x}_1=\alpha x_1-\beta x_1 x_2 \,,\\
\dot{x}_2=\delta x_1 x_2-\gamma x_2 \,, 
\end{cases}
\end{equation}
where $x_1$ and $x_2$ denote the population states of two species --- predator and prey, respectively. The parameters $\alpha$ and $\gamma$ govern the growth rate of species, while $\beta$ and $\delta$ characterize the influence of the opponent species. In this example, the system parameters are set to $[\alpha, \beta, \delta, \gamma] = [1.5,1,1,3]$. 

For data generation, the dynamical system is solved using the implicit Euler method with time step $\Delta t =0.001$ over $[0,T]$ where $T= 20$, and the initial condition is $x_1(0) = x_2(0) = 1$. The generated data are split into two phases, $[0, 0.4T]$ for training and $(0.4T, T]$ for testing/prediction. While the step size $\Delta t$ is chosen to ensure the stability and accuracy of time integration, such data density is far beyond needed for model training. We consider that only $25\%$ randomly sampled data from those collected over $[0, 0.4T]$ are used for training. This subset is treated as the entire dataset available, i.e., the case of $100\%$ data (2000 points) in the upcoming results. To investigate the impact of data sparsity on parameter estimation and uncertainty quantification, we further reduce the data density to $10\%$, $5\%$, and even $1\%$, (i.e., 200, 100 and 20 data points, respectively). Meanwhile, we introduce three levels of white noise -- $0\%, 10\%$, and $20\%$, respectively -- to the training data. The noise term follows a normal distribution with zero mean, and its standard deviation is defined by multiplying the average of data by the noise level.

We are going to test the two cases in Scenario (I) introduced in Subsection 3.1. When analyzing the results of parameter estimation, the overall relative error in the posterior means of the estimated parameters and their relative deviation, respectively denoted by $\epsilon_1$ and $\epsilon_2$, are defined as follows: 
\begin{equation}\label{eq:relatives}
    \epsilon_1= 100\% \times \left(\frac{\sum_{i=1}^{2}\|\vb*{\theta}^*_i -\mathbb{E}_\text{post}[\vb*{\theta}_i]\|_2^2}{\sum_{i=1}^{2}\|\vb*{\theta}^*_i\|_2^2}\right)^{1/2}\,,\,\text{and}\,
    \epsilon_2 = 100\% \times \left(\frac{\sum_{i=1}^{2}\mathbb{V}ar_\text{post}[\vb*{\theta}_i]}{\sum_{i=1}^{2}\|\vb*{\theta}^*_i\|_2^2}\right)^{1/2}\,,
\end{equation}
in which $\mathbb{E}_\text{post}[\vb*{\theta}_i]$ and $\mathbb{V}ar_\text{post}[\vb*{\theta}_i]$ respectively represent the posterior means and variances of parameters $\vb*{\theta}_i$, and $\vb*{\theta}^*_i$ is the ground truth (reference solution).

\begin{table}[tb]
\centering
\begin{adjustbox}{width=1\textwidth}
\begin{tabular}{c|cccc|c}
\toprule
& \multicolumn{5}{c}{noise free}    \\
\midrule
\multicolumn{1}{c|}{parameters} 
& \multicolumn{1}{c|}{100\% data}  & 
\multicolumn{1}{c|}{10\% data} &
\multicolumn{1}{c|}{5\% data} &
\multicolumn{1}{c|}{1\% data} &
ground truth\\
\midrule
$\alpha$   & \multicolumn{1}{c|}{1.500 $\pm~(6.94 \times 10^{-6})$}  
           & \multicolumn{1}{c|}{1.500 $\pm~(2.36 \times 10^{-4})$}     
           & \multicolumn{1}{c|}{1.499 $\pm~(4.19 \times 10^{-4})$}    
           & 1.511 $\pm~(2.81 \times 10^{-3})$
           & 1.5
           \\ \midrule
$\beta$    & \multicolumn{1}{c|}{1.001 $\pm~(4.29 \times 10^{-5})$}      
           & \multicolumn{1}{c|}{1.001 $\pm~(1.47 \times 10^{-4})$}     
           & \multicolumn{1}{c|}{1.000 $\pm~(3.03 \times 10^{-4})$}    
           & 1.022 $\pm~(5.83 \times 10^{-3})$
           & 1
            \\\midrule
$\delta$   & \multicolumn{1}{c|}{1.001 $\pm~(6.11 \times 10^{-5})$}      
           & \multicolumn{1}{c|}{1.001 $\pm~(2.10 \times 10^{-4})$}     
           & \multicolumn{1}{c|}{1.000 $\pm~(3.74 \times 10^{-4})$}    
           &  0.975 $\pm~(5.27 \times 10^{-3})$
           & 1
            \\\midrule
$\gamma$   & \multicolumn{1}{c|}{3.002 $\pm~(1.90 \times 10^{-4})$}      
           & \multicolumn{1}{c|}{3.003 $\pm~(6.41 \times 10^{-4})$}     
           & \multicolumn{1}{c|}{2.999 $\pm~(1.16 \times 10^{-3})$}    
           & 2.964 $\pm~(8.42 \times 10^{-3})$
           & 3
            \\
\midrule
& \multicolumn{5}{c}{$10\%$ noise level}    \\
\midrule
\multicolumn{1}{c|}{parameters} 
& \multicolumn{1}{c|}{100\% data}  & 
\multicolumn{1}{c|}{10\% data} &
\multicolumn{1}{c|}{5\% data} &
\multicolumn{1}{c|}{1\% data} &
ground truth\\
\midrule
$\alpha$   & \multicolumn{1}{c|}{1.506 $\pm~(6.51 \times 10^{-3})$}  
           & \multicolumn{1}{c|}{1.486 $\pm~(2.21 \times 10^{-2})$}     
           & \multicolumn{1}{c|}{1.449 $\pm~(3.20 \times 10^{-2})$}    
           & 1.579 $\pm~(1.25 \times 10^{-1})$
           & 1.5
           \\\midrule
$\beta$    & \multicolumn{1}{c|}{1.003 $\pm~(4.03 \times 10^{-3})$}      
           & \multicolumn{1}{c|}{0.979 $\pm~(1.37 \times 10^{-2})$}     
           & \multicolumn{1}{c|}{0.937 $\pm~(1.97 \times 10^{-2})$}    
           & 1.226 $\pm~(1.16 \times 10^{-1})$
           & 1
            \\\midrule
$\delta$   & \multicolumn{1}{c|}{1.002 $\pm~(2.82 \times 10^{-3})$}      
           & \multicolumn{1}{c|}{0.996 $\pm~(8.50 \times 10^{-3})$}     
           & \multicolumn{1}{c|}{0.990 $\pm~(1.35 \times 10^{-2})$}    
           & 0.792 $\pm~(6.45 \times 10^{-2})$
           & 1
            \\\midrule
$\gamma$   & \multicolumn{1}{c|}{2.991 $\pm~(8.74 \times 10^{-3})$}      
           & \multicolumn{1}{c|}{2.992 $\pm~(2.66 \times 10^{-2})$}     
           & \multicolumn{1}{c|}{2.967 $\pm~(4.25 \times 10^{-2})$}    
           & 2.322 $\pm~(1.65 \times 10^{-1})$
           & 3
            \\
\midrule
& \multicolumn{5}{c}{$20\%$ noise level}    \\
\midrule
\multicolumn{1}{c|}{parameters} 
& \multicolumn{1}{c|}{100\% data}  & 
\multicolumn{1}{c|}{10\% data} &
\multicolumn{1}{c|}{5\% data} &
\multicolumn{1}{c|}{1\% data} &
ground truth\\
\midrule
$\alpha$   & \multicolumn{1}{c|}{1.509 $\pm~(1.31 \times 10^{-2})$}  
           & \multicolumn{1}{c|}{1.474 $\pm~(4.45 \times 10^{-2})$}     
           & \multicolumn{1}{c|}{1.389 $\pm~(6.28 \times 10^{-2})$}    
           & 1.593 $\pm~(2.13 \times 10^{-1})$
           & 1.5
           \\\midrule
$\beta$    & \multicolumn{1}{c|}{1.001 $\pm~(8.06 \times 10^{-3})$}      
           & \multicolumn{1}{c|}{0.956 $\pm~(2.70 \times 10^{-2})$}     
           & \multicolumn{1}{c|}{0.881 $\pm~(3.79 \times 10^{-2})$}    
           & 1.304 $\pm~(1.91 \times 10^{-1})$
           & 1
            \\\midrule
$\delta$   & \multicolumn{1}{c|}{1.004 $\pm~(5.65 \times 10^{-3})$}      
           & \multicolumn{1}{c|}{0.976 $\pm~(1.72 \times 10^{-2})$}     
           & \multicolumn{1}{c|}{0.986 $\pm~(2.74 \times 10^{-2})$}    
           &  0.758 $\pm~(1.17 \times 10^{-1})$
           & 1
            \\\midrule
$\gamma$   & \multicolumn{1}{c|}{2.981 $\pm~(1.74 \times 10^{-2})$}      
           & \multicolumn{1}{c|}{2.933 $\pm~(5.38 \times 10^{-2})$}     
           & \multicolumn{1}{c|}{2.970 $\pm~(8.57 \times 10^{-2})$}    
           & 2.050 $\pm~(2.99 \times 10^{-1})$
           & 3
            \\
\bottomrule

\end{tabular}
\end{adjustbox}
\caption{Mean $\pm$ standard deviation of estimated model parameters under 0\%, 10\% and 20\% noise level with 100\%, 10\%, 5\%, and 1\% data densities (2000, 200, 100, and 20 data points, respectively)}. 
\label{tab:OpInf_parameters}
\end{table}

\begin{figure}[th!]
  \centering
  \includegraphics[width=0.72\textwidth]{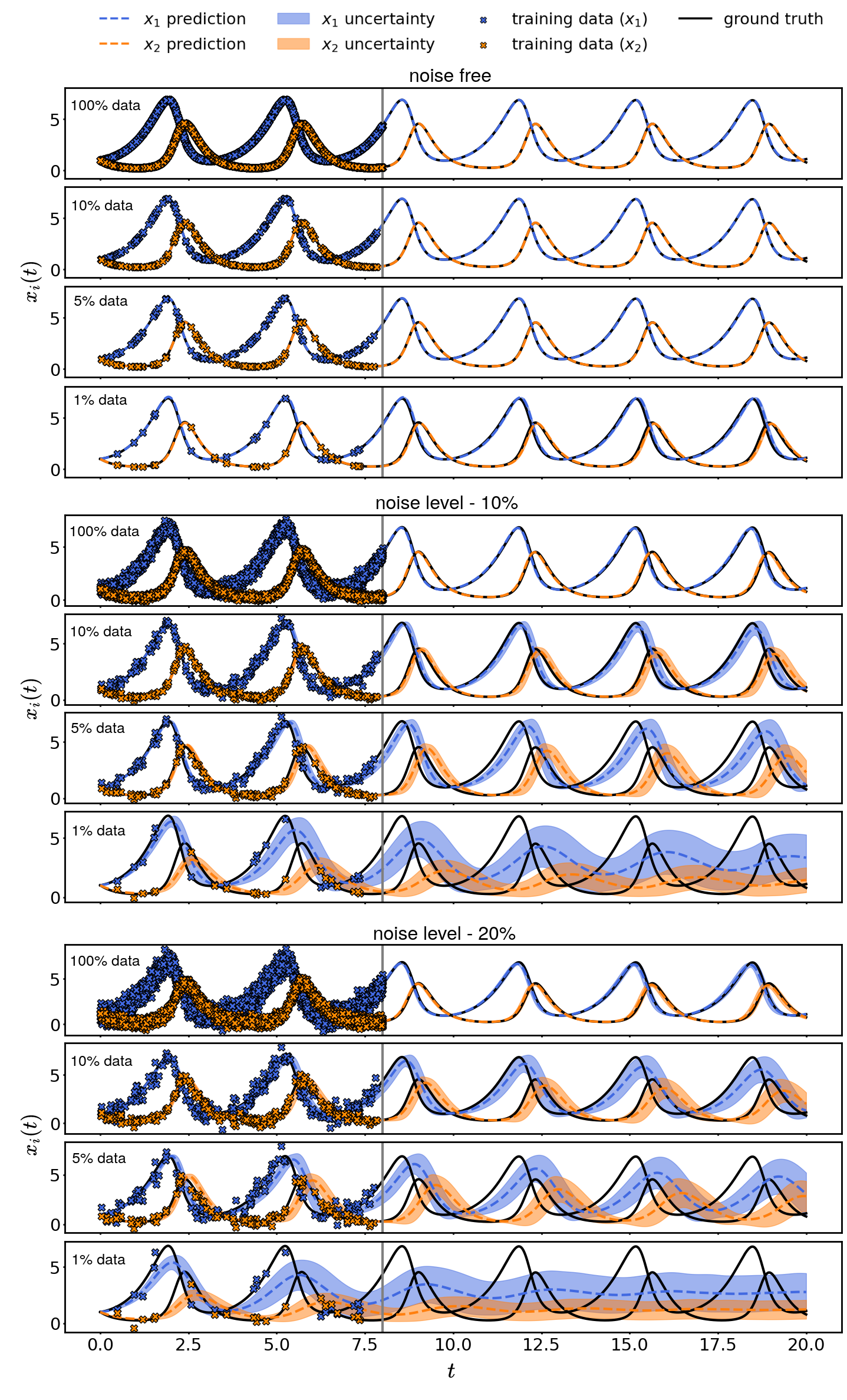}
  \caption{Trajectory reconstruction ($t\in [0,8]$) and prediction ($t\in(8,20]$) of prey and predator dynamics with estimated parameters at $0\%$, $10\%$, and $20\%$ noise levels using $100\%$, $10\%$, $5\%$, and $1\%$ data (2000, 200, 100, and 20 data points, respectively). The uncertainty bands denote $\pm~\sigma$ (standard deviation) of the Bayesian prediction. The grey vertical line separates the training and testing stages.}
\label{fig:sc1_prediction}
\end{figure}

\subsubsection*{Case A: parameter estimation}

In Case A, we take
\begin{equation*}
    \begin{split}
        & \vb*{g}_1(\vb*{x}) = [~ x_1 ~~ x_1 x_2~]\trp\,,\quad \vb*{\theta}_1 = [~\alpha ~~ -\beta~]\trp\,;\\
        & \vb*{g}_2(\vb*{x}) = [~x_1 x_2 ~~ x_2~]\trp\,,\quad \vb*{\theta}_2 = [~\delta ~~ -\gamma~]\trp\,,
    \end{split}
\end{equation*}
and thus the number of parameters for each equation is $p_i=2$, $i=1,2$. Here we incorporate no prior assumption on $\vb*{\theta}_i$ by taking $\lambda_i  = 0$, i.e., the prior distribution is uninformative and the corresponding Tikhonov regularization in \eqref{eq:L2posterior} vanishes.

The posterior mean and standard deviation of $\vb*{\theta}_1$ and $\vb*{\theta}_2$ estimated under various conditions are presented in Table~\ref{tab:OpInf_parameters}. In the noise-free cases, the inferred parameters coincide with the ground truth, and the small standard deviations indicate low uncertainties in the estimation. With less training data, a minor reduction in the accuracy of the mean predictions is observed, accompanied by naturally expanded deviations. Nevertheless, even when utilizing only $1\%$ of the data, the inference result still aligns closely with the ground truth, and the corresponding uncertainty remains at the $10^{-3}$ level. In the noise-corrupted cases, the compromise of accuracy in parameter inference becomes more obvious, especially when the data density is diminished. Although the posterior mean values remain accurate with $100\%$ data, the standard deviations are increased by the corruption of data noise. The most challenging case arises when $1\%$ of the data with $20\%$ noise level are adopted. It leads to a complete breakdown of the parameter estimation. 

\begin{figure}[t!]
  \centering
  \includegraphics[width=\textwidth]{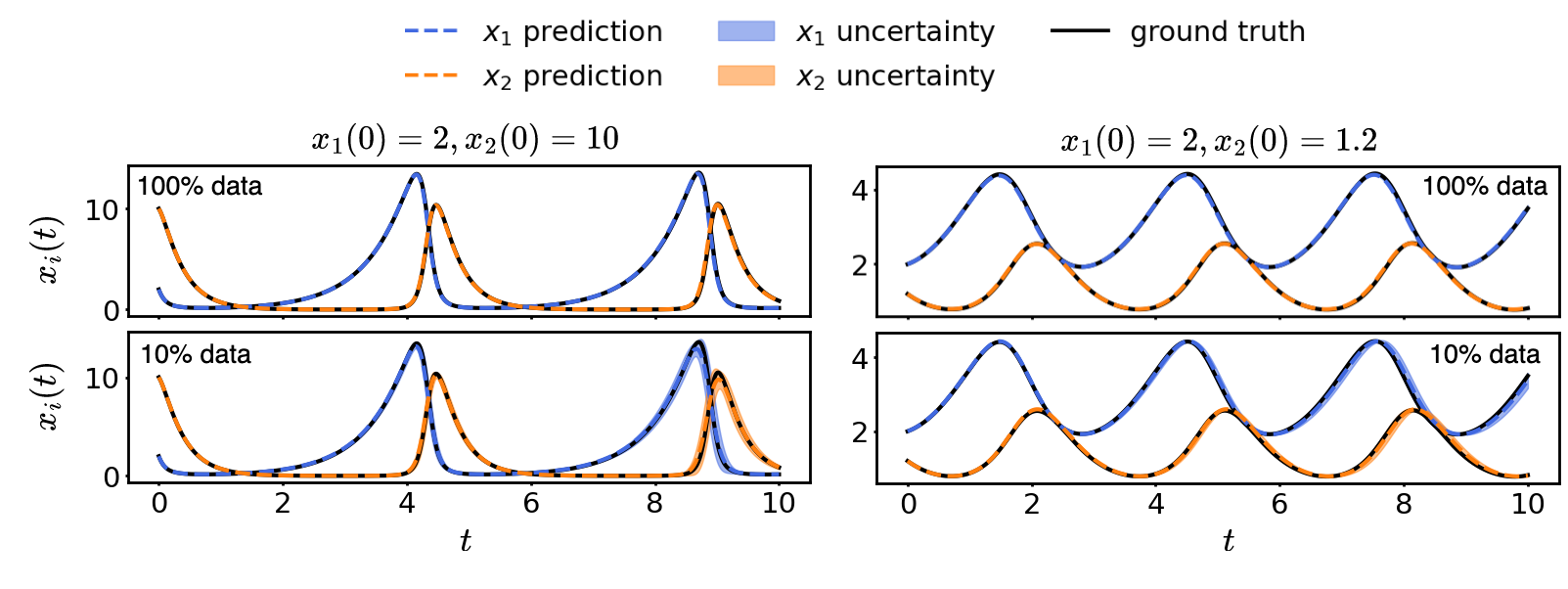}
  \caption{Trajectory predictions for two new initial conditions issued by dynamical models respectively trained with $100\%$ and $10\%$ of data (2000 and 200 data points, respectively), both corrupted by $10\%$ noise.}
\label{fig:sc1_prediction_ic}
\end{figure}

The obtained posterior distributions of system parameters are subsequently employed in Bayesian prediction using \ref{eq:Bayes_prediction}. The predictions are visualized in Figure~\ref{fig:sc1_prediction}. The time window before the vertical gray line ($t=0.4T=8$) corresponds to the training data coverage, while the window afterwards is the prediction phase. Evidently, the existence of data noise significantly impacts the predictive uncertainty and requires a sufficient amount of data to guarantee accuracy. 
Solutions under two new initial conditions, respectively being $(x_1(0),x_2(0)) = (2,10)$ and $(x_1(0),x_2(0))=(2,1.2)$, are evaluated numerically and shown in Figure~\ref{fig:sc1_prediction_ic}. These two examples serve to demonstrate the fact that the learned dynamical model is (and should be) independent of the initial conditions of training trajectories. 
These solutions are respectively derived with $100\%$ and $10\%$ training data densities, both corrupted by $10\%$-level noise. The results present a close alignment between the predictions and  ground truth of the trajectories.

\begin{figure}[th]
  \centering
  \includegraphics[width=\textwidth]{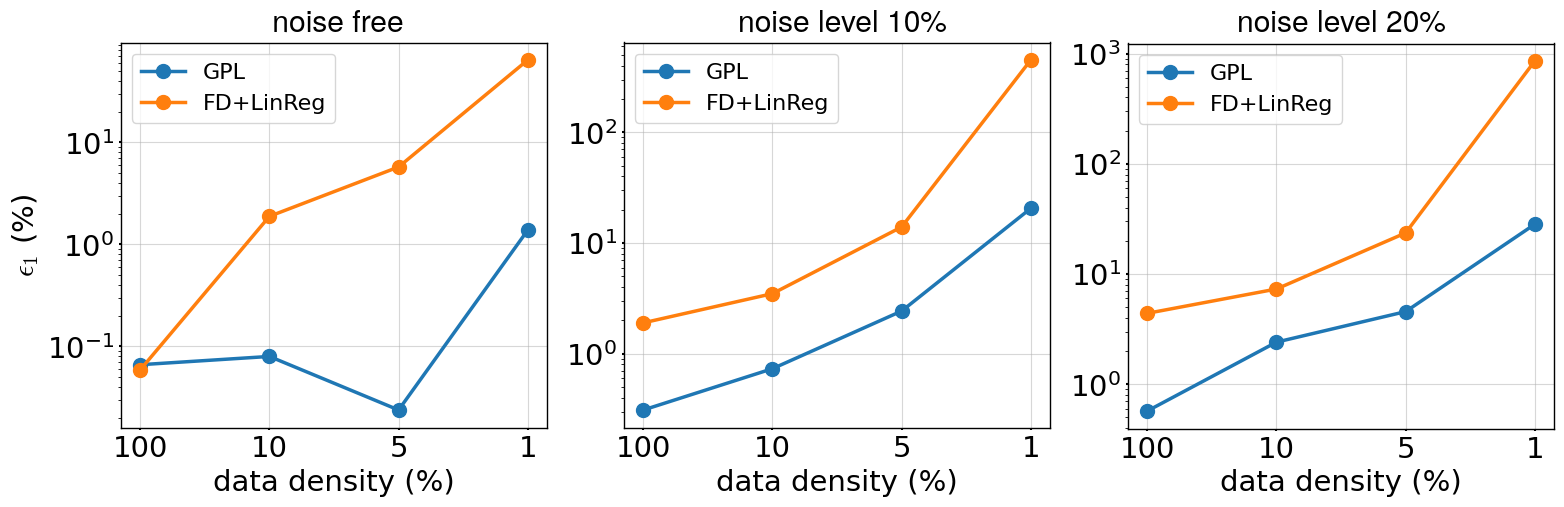}
  \caption{Comparison of relative errors in the estimated parameters by the proposed Gaussian process learning (GPL) and those by linear regression with finite-difference derivative approximations (FD+LinReg). Both models are trained with 100\%, 10\%, 5\% and 1\% data (2000, 200, 100, and 20 points, respectively) under 0\%, 10\%, and 20\% noise levels. In the latter case, noisy data are smoothed by the Savitzky-Golay method \cite{gorry1990general}.}
\label{fig:sc1_prediction_compare}
\end{figure}

To show the advantage of the proposed method, a comparison of relative errors in estimated parameters by Gaussian process learning and those by linear regression with finite-difference derivative approximation is depicted in Figure~\ref{fig:sc1_prediction_compare}. The relative error $\epsilon_1$ was defined in \eqref{eq:relatives}. Both models are trained with 100\%, 10\%, 5\%, and 1\% data under 0\%, 10\% and 20\% noise levels. Here, noisy data are smoothed by the Savitzky-Golay method \cite{gorry1990general}. Though the linear regression with finite-difference derivatives exhibits comparable accuracy to the proposed method when 100\% noise-free data are used for training, it is consistently outperformed by the proposed method when noise is added or data are reduced.


\subsubsection*{Case B: sparse identification}

\begin{figure}[th!]
  \centering
  \includegraphics[width=.95\textwidth]{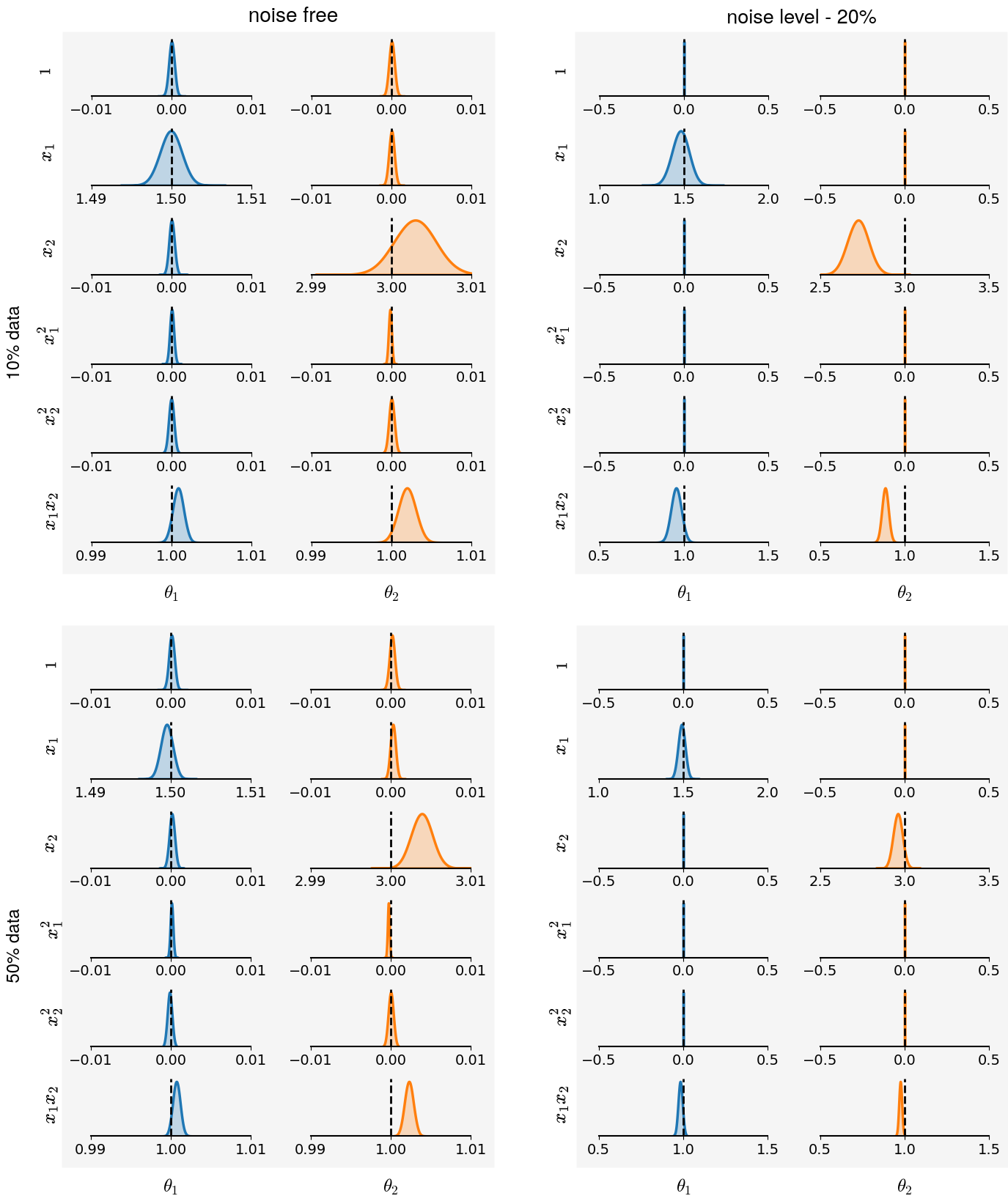}
  \caption{Posterior distributions of $\vb*{\theta}_i$ estimated under 0\% and 20\% noise levels with 10\% and 50\% data (200 and 1000 data points, respectively).}
\label{fig:sc2_para}
\end{figure}

\begin{figure}[th!]
  \centering
  \includegraphics[width=0.95\textwidth]{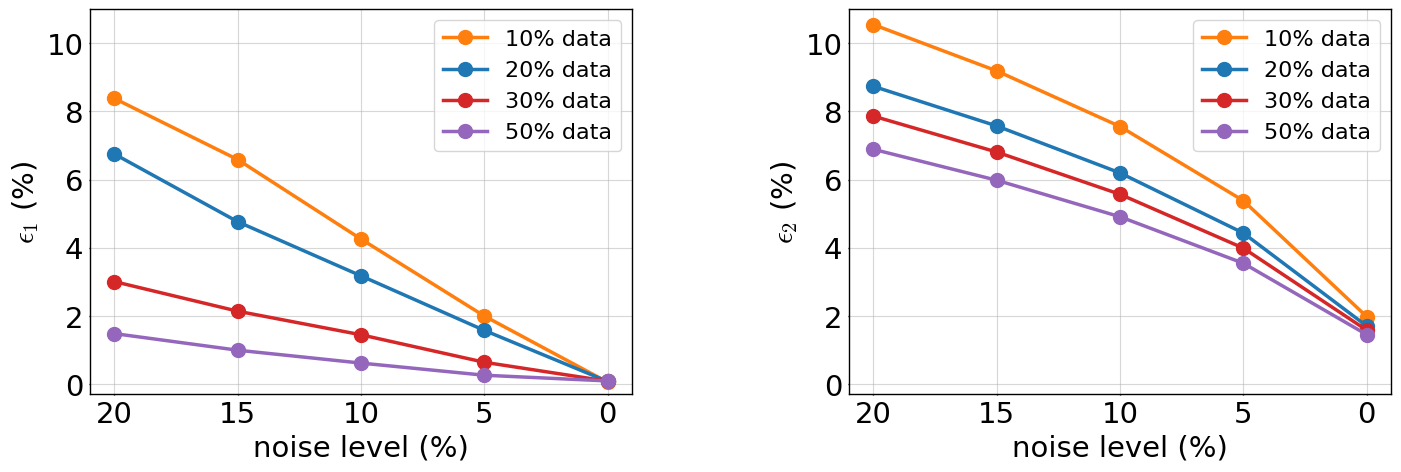}
  \caption{Relative error of mean estimates ($\epsilon_1$) and relative deviation ($\epsilon_1$) in the posterior distributions of parameters with various noise levels and data densities.}
\label{fig:sc2_compare}
\end{figure}

\begin{figure}[th!]
  \centering
  \includegraphics[width=\textwidth]{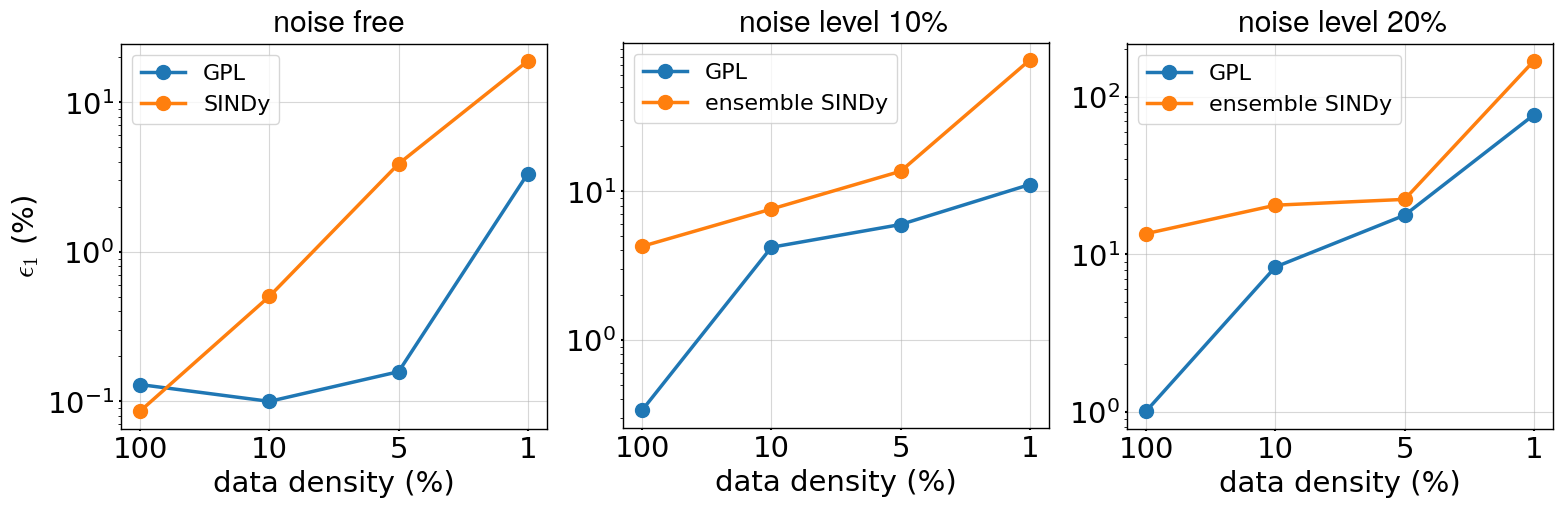}
  \caption{Comparison of relative errors in the estimated parameters by the proposed Gaussian process learning (GPL) and those by SINDy. Both models are trained with 100\%, 10\%, 5\%, and 1\% data (2000, 200, 100, and 20 data points, respectively) under 0\%, 10\%, and 20\% noise levels. In the noisy cases, the ensemble SINDy method with a bragging scheme is employed \cite{Fasel2022}.}
\label{fig:Sindy}
\end{figure}

In this case, we set
\begin{equation*}
    \begin{split}
        & \vb*{g}_1(\vb*{x}) = [~1 ~~x_1 ~~x_2 ~~x_1^2 ~~x_2^2 ~~x_1 x_2~]\trp\,,\quad \vb*{\theta}_1 = [~0 ~~\alpha ~~0 ~~0 ~~0 ~~-\beta~]\trp\,;\\
        & \vb*{g}_2(\vb*{x}) =[~1 ~~x_1 ~~x_2 ~~x_1^2 ~~x_2^2 ~~x_1 x_2~]\trp\,,\quad \vb*{\theta}_2 = [~0 ~~0 ~~-\gamma ~~0 ~~0 ~~\delta ~]\trp\,,
    \end{split}
\end{equation*}
where $p_i=6$, $i=1,2$, and only two out of six terms in each $\vb*{g}_i$ are active. A sparse identification of these active terms is thus needed together with the inference of parameters. Once the sparsity is recognized through the sequential threshold regression \eqref{eq:sqtrr}, the hyperparameters $\lambda_i^\text{s}$ and $\lambda_i^\text{a}$ are chosen to be $10^7$ and $10^{-7}$, respectively, to enforce the sparsity promotion in the prior distribution. Similar to Case A, results of parameter estimation with different noise levels and amounts of data are presented. The posterior distributions of $\vb*{\theta}_1$ and $\vb*{\theta}_2$ with 10\%, 50\% data corrupted by 0\%, 20\% noise are depicted in Figure~\ref{fig:sc2_para}. In the noise-free cases, both 10\% and 50\% data give meaningful results of posterior distributions: those of sparsified terms are centered closely to zero and have minuscule variance values, while the active terms have their mean values close to the ground truth and deviate in a wider range. When the amount of data is increased to 50\%, the variances of inferred parameters are effectively reduced consequently. When 20\% noise is added to data, there is clearly an overall increase in posterior variances, and the accuracy of mean estimates is slightly compromised compared to their noise-free counterparts. 
All these observations align with the intuition of uncertainty quantification. 

The relative error in the posterior means and the relative deviation of the estimated parameters at different noise levels and amounts of training data, as previously defined in \eqref{eq:relatives}, are shown in Figure~\ref{fig:sc2_compare}. It is evident that, overall, the values of both these error/uncertainty indicators are effectively decreased as the data amount is increased or the noise is reduced, i.e., an convergence is confirmed by these observations.

Besides, we present a comparison of relative errors between the proposed Gaussian process learning and SINDy, as shown in Figure~\ref{fig:sc2_compare}. Both models are trained with 100\%, 10\%, 5\%, and 1\% data under 0\%, 10\% and 20\% noise levels. In the noisy cases, ensemble SINDy with a bragging scheme is employed \cite{Fasel2022} to improve the performance of SINDy under uncertainties. We have observed that the proposed Gaussian-process-based method outperformed the SINDy methods in terms of estimation accuracy, particularly when dealing with noisy and sparse data.

\subsection{1D nonlinear ordinary differential equation}

\begin{figure}[t!]
  \centering
  \includegraphics[width=0.85\textwidth]{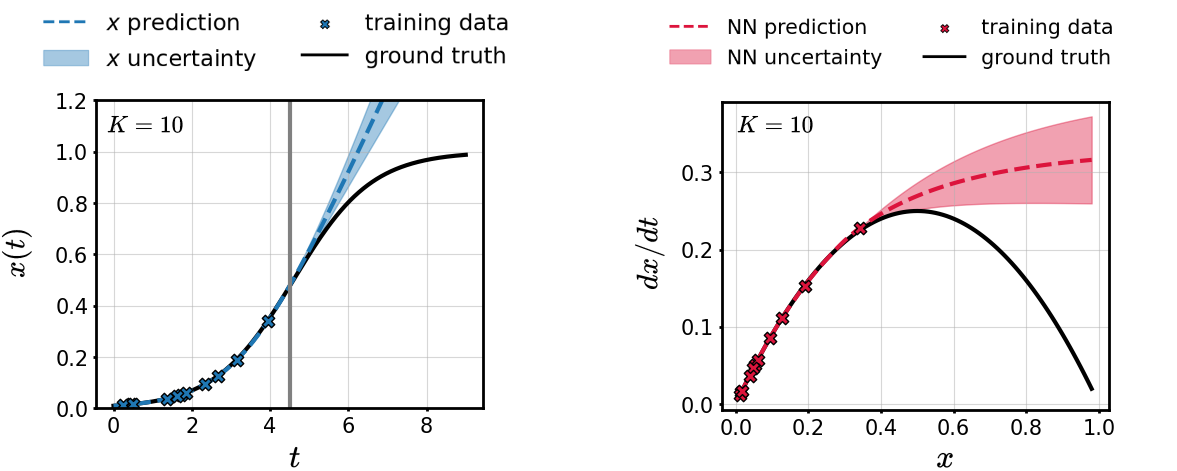}
  \caption{(Left) Trajectory reconstruction ($t\in [0,4.5]$) and prediction ($t\in(4.5,9]$) by a shallow neural network trained with 10 noise-free data points. The uncertainty bands denote $\pm~\sigma$ (standard deviation) of the Bayesian prediction. The grey vertical line separates the training and testing stages. (Right) Corresponding neural network approximation of $f(\cdot,\vb*{\theta})$. The uncertainty bands denote $\pm~\sigma$ (standard deviation) of $f(\cdot,\vb*{\theta})$ approximation. }
\label{fig:sc3_approxi}
\end{figure}
\begin{figure}[h!]
  \centering
  \includegraphics[width=0.85\textwidth]{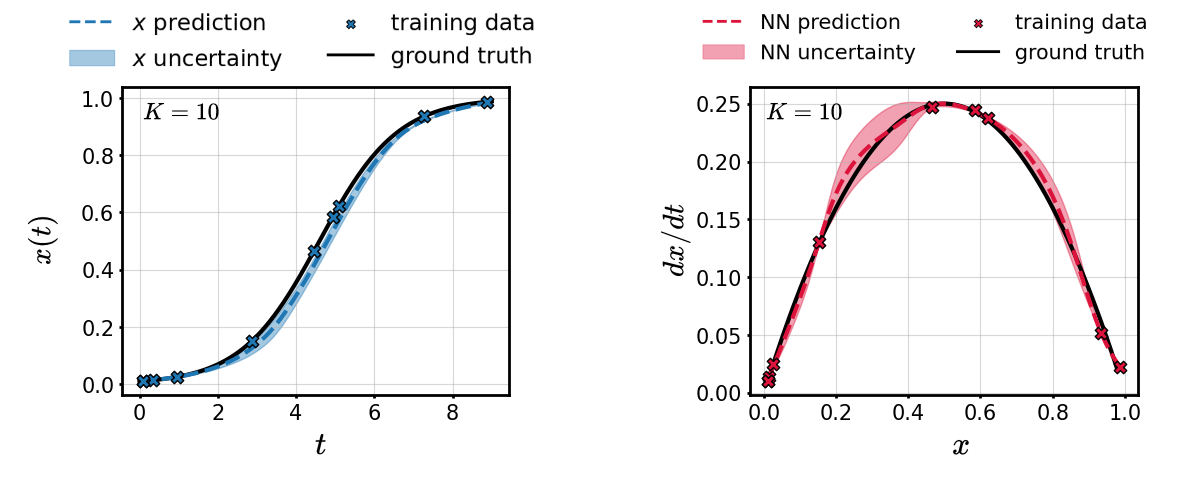}
  \caption{(Left) Trajectory reconstruction ($t\in [0,9]$) by a shallow neural network trained with 10 noise-free data points. The uncertainty bands denote $\pm~\sigma$ (standard deviation) of the Bayesian prediction. (Right) Neural network approximation of $f(\cdot,\vb*{\theta})$. The uncertainty bands denote $\pm~\sigma$ (standard deviation) of $f(\cdot,\vb*{\theta})$ approximation.}
\label{fig:sc3_approxi_full}
\end{figure}

\begin{figure}[h]
  \centering
  \includegraphics[width=0.85\textwidth]{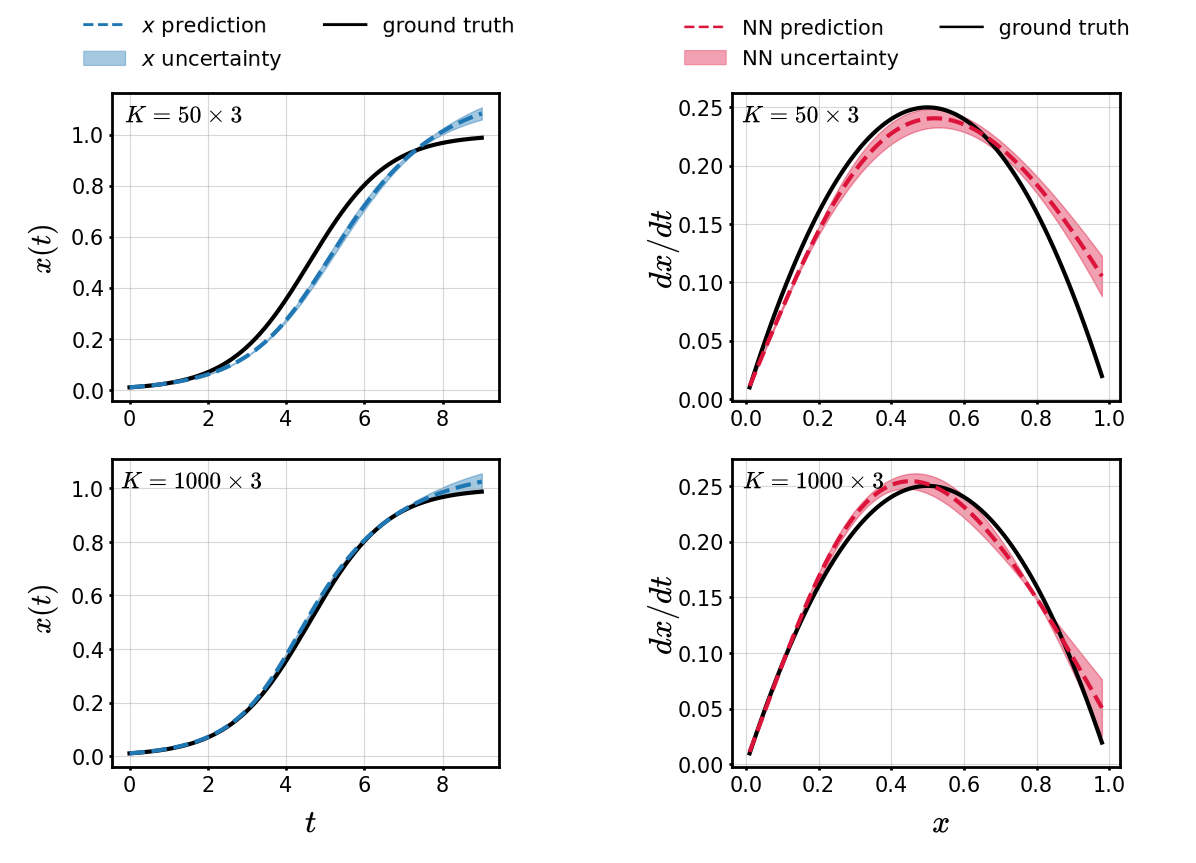}
  \caption{(Left column) Trajectory reconstruction ($t\in [0,9]$) by a shallow neural network trained with data from three initial conditions $\gamma=0.01, 0.2$ and $0.7$. The data is corrupted by 10\% noise. The uncertainty bands denote $\pm~\sigma$ (standard deviation) of the Bayesian prediction. (Right column) Corresponding neural network approximation of $f(\cdot,\vb*{\theta})$ . The uncertainty bands denote $\pm~\sigma$ (standard deviation) of $f(\cdot,\vb*{\theta})$ approximation.}
\label{fig:sc3_noise}
\end{figure}

We present an example of 1D nonlinear ordinary differential equation ($N=1$) to demonstrate the aforementioned Scenario (II), in which the dynamical model to be learned is approximately parameterized by a shallow neural network. Training data are generated from a reference model equation as follows:
\begin{equation}\label{eq:nn}
    \dot{x} = x(1-x)\,,\quad x(0) = \gamma\,,
\end{equation}
in which $\gamma$ specifies the initial condition. This initial-value problem has an analytical solution 
\begin{equation}
    x(t) = \frac{\gamma}{\gamma+(1-\gamma)e^{-t}}\,,\quad t > 0\,.
\end{equation}
We use this simple example to show the limitation of neural network approximation, as well as the proposed method's robust performance in uncertainty quantification. For data generation, the analytical solution is evaluated over $t\in [0,T]$ with $T=9$ under a given initial condition.

Similar to the previous example, we first divide the trajectory into training $[0,0.5T]$ and testing/prediction $(0.5T,T]$ stages. 10 noise-free data points randomly sampled over $[0, 0.5T]$ with the initial condition $\gamma = 0.01$ are used for training. $f(\cdot;\vb*{\theta})$\footnote{The subscript of $\vb*{\theta}_i$ is omitted here as $i=1$ only.} is approximated by a shallow neural network with a single hidden layer of 8 neurons and hyperbolic tangent activation, and all the network parameters (i.e., weights and biases) are collected in the vector $\vb*{\theta}$. Independent Gaussian prior with zero mean and variance $100$ ($\lambda = 0.01$) is assigned to each parameter. 

Results with scarce, noise-free data, including the trajectory reconstruction and prediction, as well as the approximation of $f(\cdot;\vb*{\theta})$, are presented in Figure~\ref{fig:sc3_approxi}. The neural network approximates $f(\cdot;\vb*{\theta})$ well within the range covered by training data, but fails in extrapolation when the prediction of $x$ goes beyond the training coverage. In fact, this observation is aligned with the limitation of neural network regression in generalization. In contrast to the previously discussed scenarios where the dictionary $\boldsymbol{g}_i(\boldsymbol{x})$ of candidate terms are considered to cover the expression of $f_i(\boldsymbol{x};\cdot)$, the neural network ansatz space here does not exactly include the actual underlying system representation but provides a reparametrization/approximation. This is commonly considered in the cases with unspecified nonlinear parametrization. Therefore, the poor predictive performance beyond training coverage is reasonable and expected. Nevertheless, such inadequacy in predictions can be effectively indicated by the uncertainty bands in our results. On the other hand, if the training data have covered the entire time domain of interest, the neural network reparametrization $f(\cdot;\vb*{\theta})$ serves a surrogate model. As shown in Figure~\ref{fig:sc3_approxi_full}, only with 10 data points, the surrogate modeling is already decent and enables a good trajectory reconstruction.

As discussed in Remark 2, one way to improve the approximation is to incorporate the data generated from multiple initial conditions for training. Such multiple sets of data explore different ranges of $x$ and subsequently extends the training coverage for $f(\cdot;\vb*{\theta})$, which also makes the dynamics learning less dependent on a single initial condition. In light of this, we consider including three initial conditions: $\gamma=0.01$, $0.2$ and $0.7$. For each initial condition, training data are sampled over $t\in[0,0.2T]$, and two experiments respectively with 50 and 1000 data points are performed. Moreover, each data point is corrupted by an independent noise term that follows a normal distribution with zero mean and a standard deviation defined by multiplying the noise level -- 10\%. In this case, the results of neural network approximation and trajectory reconstruction are depicted in Figure~\ref{fig:sc3_noise}. It can be observed that, despite being impacted by the added noise, the proposed method still manages to capture the underlying dynamics. With an increase in the data amount, the approximation of $f(\cdot;\vb*{\theta})$ is improved and hence leads to a more accurate trajectory reconstruction. 

\subsection{Orbital dynamics of black hole} 

\begin{figure}[t!]
  \centering
  \includegraphics[width=0.83\textwidth]{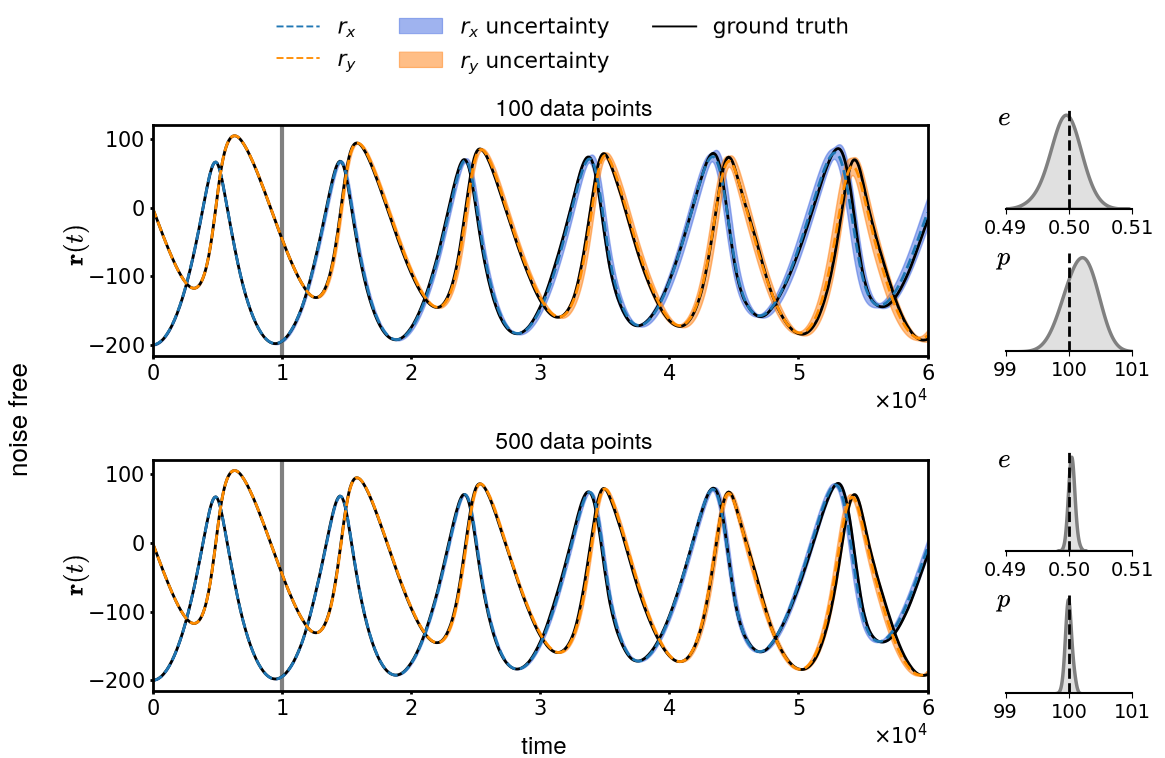}
  \caption{(Left) Trajectory reconstruction ($t\in [0,1]\times 10^4$) and prediction ($t\in (1,6]\times 10^4$) of black hole orbital dynamics with estimated parameters using 100 and 500 noise-free data points. The uncertainty bands denote $\pm~\sigma$ (standard deviation) of the Bayesian prediction. (Right) Posterior distributions of the inferred parameters $e$ and $p$.}
\label{fig:sc4_noisefree}
\end{figure}

\begin{figure}[h!]
  \centering
  \includegraphics[width=0.83\textwidth]{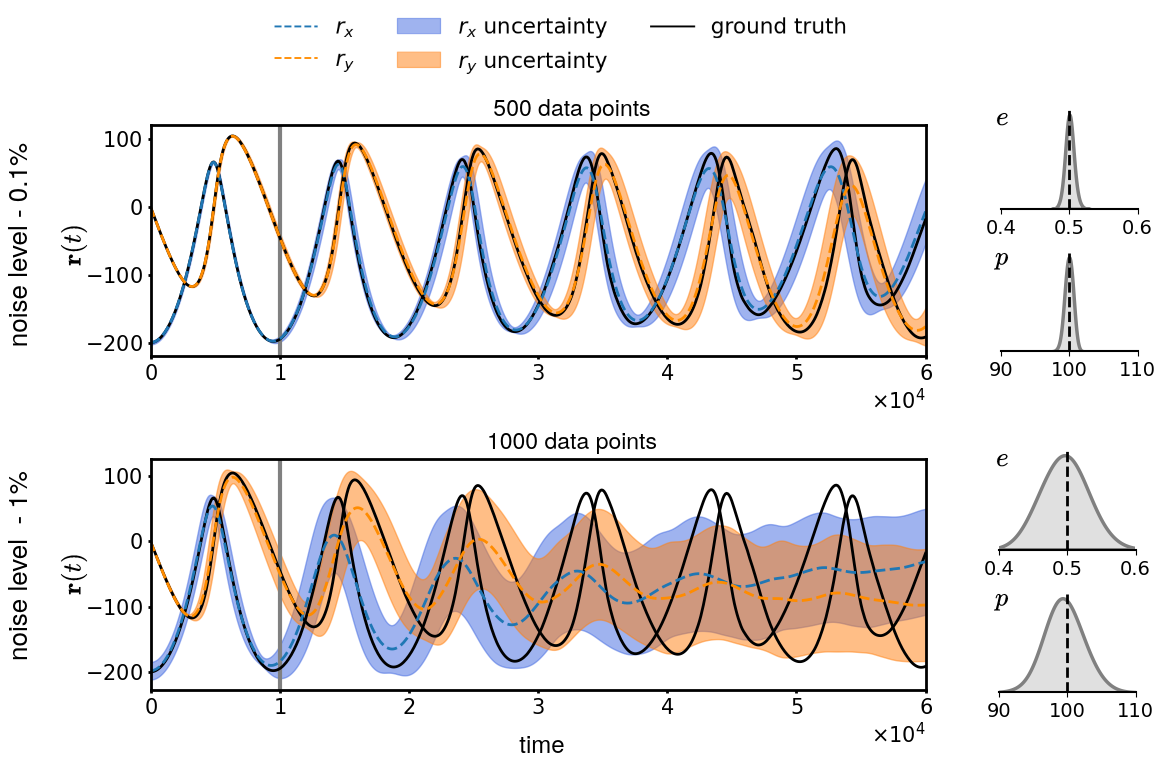}
  \caption{(Left) Trajectory reconstruction ($t\in [0,1]\times 10^4$) and prediction ($t\in (1,6]\times 10^4$) of black hole orbital dynamics with estimated parameters using 500 data points under 0.1\% noise level and 1000 data points under 1\% noise level. The uncertainty bands denote $\pm~\sigma$ (standard deviation) of the Bayesian prediction. (Right) Posterior distributions of the inferred parameters $e$ and $p$.}
\label{fig:sc4_noise}
\end{figure}

\begin{figure}[th!]
  \centering
  \includegraphics[width=0.92\textwidth]{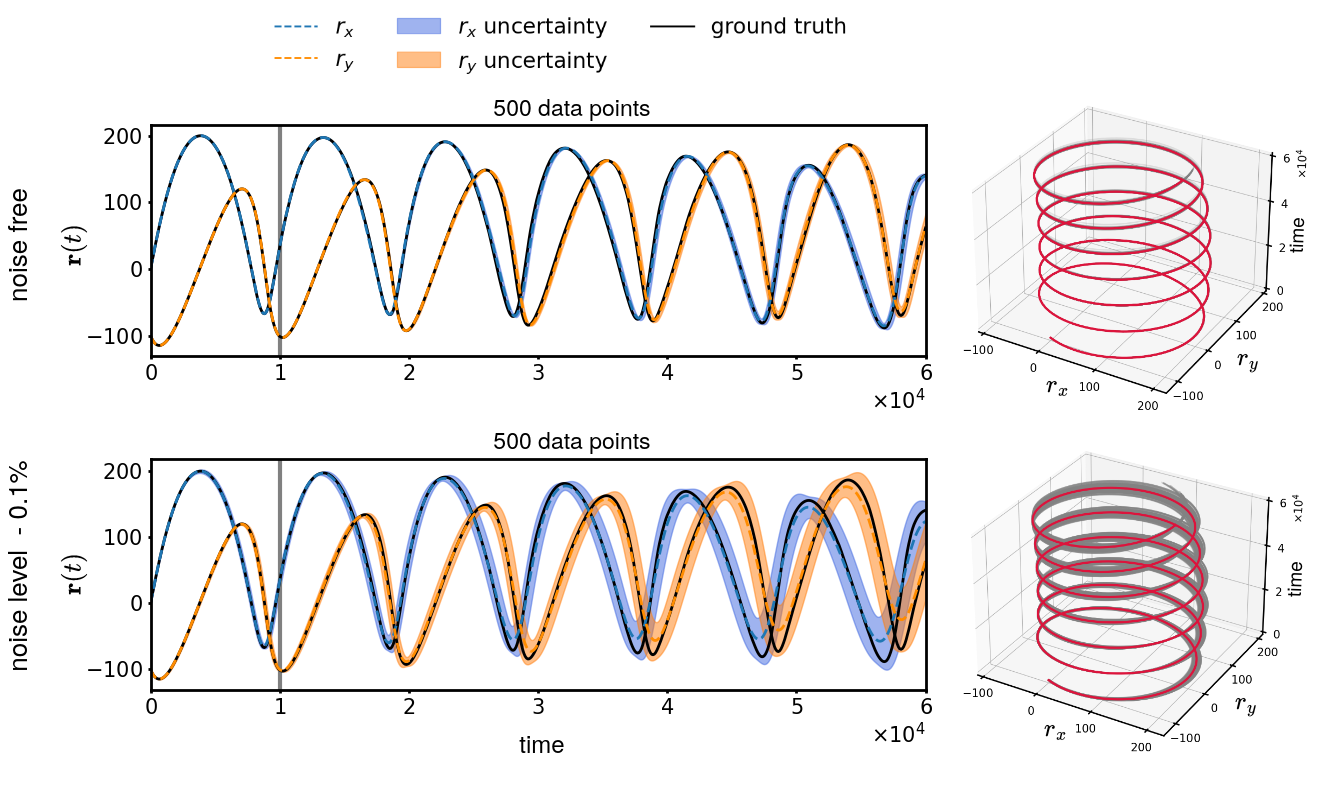}
  \caption{(Left) Trajectory predictions for a new initial condition $\phi(0) = \chi(0)=\frac{\pi}{2}$. The parameters are estimated with 500 data points under 0\% and 0.1\% noise level. The uncertainty bands denote $\pm~\sigma$ (standard deviation) of the Bayesian prediction. (Right) 3-D view of the trajectories, in which the red curve shows the ground truth, and the grey ones represent the ensemble of trajectories evaluated with the sampled parameter values from the posterior distributions.}
\label{fig:sc4_IC}
\end{figure}

In this example, the proposed method is applied to general parameter estimation for a given nonlinear dynamical system. We consider the binary black hole model \cite{chandrasekhar1998mathematical} that describes the relativistic orbital dynamics of a two-body problem, and particularly focus on a simplified case where one object in the system has a significantly greater mass than the other. In this case, the more massive one stays still at the origin of the coordinates. The trajectory of the less massive one (often referred to as the `test body') in a 2D Cartesian coordinate system can be written as
\begin{equation}
    \vb{r}{(t)} = [r_x(t)~,~r_y(t)]^{\trp} = -r(t) ~ [\cos \phi(t)~,~\sin \phi(t)]^{\trp}\,,
\end{equation}
in which $\phi$ and $r$ are respectively the angle and radius in polar notion. The radius is further given by
\begin{equation}
    r(t) = \frac{p}{1+e\cos(\chi(t))}\,,
\end{equation}
where the eccentricity $e$ and semilatus rectum $p$ (in an ellipse) are set to be constant, and the anomaly angle $\chi$ is time-dependent. Importantly, the polar angle $\phi(t)$ and anomaly $\chi(t)$ of the test body ($\vb*{x}(t)=[\phi(t),\chi(t)]\trp$) are governed by the governing equations \cite{keith2021learning}:
\begin{equation}\label{eq:blackhole}
\begin{split}
\dot{\phi} & = \frac{(p-2-2e\cos\chi)(1+e\cos\chi)^2}{p^{3/2}[(p-2)^2-4e^2]^{1/2}}\,,\\
\dot{\chi} & = \frac{(p-2-2e\cos\chi)(1+e\cos\chi)^2(p-6-2e\cos\chi)^{1/2}}{p^2[(p-2)^2-4e^2]^{1/2}} \,.   
\end{split}
\end{equation}
The task of this numerical experiment is to estimate the parameters $e$ and $p$ using our proposed model trained with time-series data of $\phi$ and $\chi$. Here $\vb*{\theta}_1 = \vb*{\theta}_2 = [e,p]\trp$, falling into the category where parameters are shared between equations (Remark 3) and \eqref{eq:shareparam} is used for Bayesian inference. The right-hand side of \eqref{eq:blackhole} defines $f_1$ and $f_2$. Specifically for the generation of training data, the dynamical system \eqref{eq:blackhole} is initialized with $\phi(0) = 0$ and $\chi(0) = \pi$, and then evaluated over $[0,T] = [0,10^4]$ with $e=0.5$ and $p=100$ using an implicit multi-step method \cite{shampine1997matlab}. 

Using the proposed method on noise-free data, we obtain the posterior distributions of $e$ and $p$ and the reconstructed/predicted trajectories shown in Figure~\ref{fig:sc4_noisefree}. The posterior mean values of the parameters are located closely to the ground truth, and the variances can be effectively reduced by increasing data density. The predicted trajectories overlap with the ground truth, confirming that the Bayesian prediction is well-performing. Thereafter, the results with 500 data points under 0.1\% noise and 1000 points under 1\% noise are depicted in Figure~\ref{fig:sc4_noise}. It is clear that the accuracy of posterior means is decent, while the variances are substantially increased. There is a rapid decay in the accuracy of the predicted trajectory under 1\% noise. This is mainly because the dynamical system is sensitive to the parameters $e$ and $p$, and thus the errors caused by imperfect parameter values propagate and accumulate fast over time. In spite of this, the parameter estimation is still reasonable and informative. In fact, the predictive mean, along with the uncertainty band, is computed from an ensemble of trajectories evaluated at the sampled parameter values from their joint posterior distribution, so the decaying predictive accuracy is also a result of the uncertainty accumulation throughout the modeling process. Notably, the $\pm \sigma$ uncertainty level basically bounds the error in the mean trajectory, indicating a meaningful uncertainty quantification. 

Furthermore, as discussed in Remark 2, parameter estimation should be independent of initial conditions, and the estimated parameters should work for new initial conditions beyond those in training data. This is confirmed again by Figure~\ref{fig:sc4_IC} that shows good results of trajectory predictions for a new initial condition $\phi(0) = \chi(0)=\frac{\pi}{2}$.

\section{Concluding remarks}

The core novelty of this work lies in defining a differential-equation constrained likelihood using Gaussian process approximation in the Bayesian parameter estimation for dynamical systems. The resulting physics-aware nature improves the interpretability of data-driven dynamics learning. Compared to existing methods with a similar purpose, the proposed method does not require a direct numerical approximation of time-derivatives from solution data, but instead involves derivative evaluations implicitly through Gaussian process emulation. This is in particular advantageous when the trajectory data are scarce and/or corrupted by noise, because the accuracy of direct derivative approximation would then be compromised considerably, while Gaussian processes are capable of achieving a good, smoothed approximation with properly chosen hyperparameters. Without involving adjoint states in the algorithmic procedure, the proposed Gaussian process likelihood can be easily implemented within a classical Bayesian framework and even used as a plug-in in many relevant methods for dynamics learning, as demonstrated in the contexts of parameter estimation, sparse identification, and neural network approximation. Importantly, the presented probabilistic treatment enables a quantification of uncertainties stemming from the modeling process and imperfect datasets, providing meaningful model validation exemplified and confirmed by numerical results.

This method provides a promising tool for data-driven discovery of governing equations, as well as for the representation of reduced-order dynamics for high-dimensional systems. Though the method should be able to produce decent estimates robustly when the amount of data is reduced to a reasonable extent, sparse Gaussian processes can be adopted alternatively to relieve the computational burden caused by large data amounts in Gaussian process emulation.

\section*{Acknowledgement}
The authors acknowledge the financial support from Sectorplan Bèta (NL) under the focus area \emph{Mathematics of Computational Science}, and would like to thank Prof. Christoph Brune for fruitful discussions.

\section*{Data availability}
All the data and source codes to reproduce the results in this study are available on GitHub at \url{https://github.com/DongweiYe/Gaussian-Process-Learning}.

\bibliographystyle{plain}
\bibliography{refs.bib}

\begin{thebibliography}{10}

\bibitem{batlle2023error}
Pau Batlle, Yifan Chen, Bamdad Hosseini, Houman Owhadi, and Andrew~M Stuart.
\newblock Error analysis of kernel/{GP} methods for nonlinear and parametric {PDE}s.
\newblock {\em arXiv preprint arXiv:2305.04962}, 2023.

\bibitem{batlle2023kernel}
Pau Batlle, Matthieu Darcy, Bamdad Hosseini, and Houman Owhadi.
\newblock Kernel methods are competitive for operator learning.
\newblock {\em arXiv preprint arXiv:2304.13202}, 2023.

\bibitem{beckers2022gaussian}
Thomas Beckers, Jacob Seidman, Paris Perdikaris, and George~J Pappas.
\newblock {G}aussian process port-{H}amiltonian systems: {B}ayesian learning with physics prior.
\newblock In {\em 2022 IEEE 61st Conference on Decision and Control (CDC)}, pages 1447--1453. IEEE, 2022.

\bibitem{Bonilla2007}
Edwin~V Bonilla, Kian Chai, and Christopher Williams.
\newblock Multi-task {G}aussian process prediction.
\newblock {\em Advances in neural information processing systems}, 20, 2007.

\bibitem{Botteghi2022}
Nicol{\`o} Botteghi, Mengwu Guo, and Christoph Brune.
\newblock Deep kernel learning of dynamical models from high-dimensional noisy data.
\newblock {\em Scientific Reports}, 12(1):21530, Dec 2022.

\bibitem{box2011bayesian}
George~EP Box and George~C Tiao.
\newblock {\em {B}ayesian Inference in Statistical Analysis}.
\newblock John Wiley \& Sons, 2011.

\bibitem{Steven2016}
Steven~L Brunton, Joshua~L Proctor, and J~Nathan Kutz.
\newblock Discovering governing equations from data by sparse identification of nonlinear dynamical systems.
\newblock {\em Proceedings of the National Academy of Sciences}, 113(15):3932--3937, 2016.

\bibitem{chandrasekhar1998mathematical}
Subrahmanyan Chandrasekhar.
\newblock {\em The Mathematical Theory of Black Holes}.
\newblock Oxford University Press, 1998.

\bibitem{Chang2015}
Eugene~TY Chang, Mark Strong, and Richard~H Clayton.
\newblock {B}ayesian sensitivity analysis of a cardiac cell model using a {G}aussian process emulator.
\newblock {\em PLOS ONE}, 10(6):1--20, 06 2015.

\bibitem{Chen2018}
Ricky~TQ Chen, Yulia Rubanova, Jesse Bettencourt, and David~K Duvenaud.
\newblock Neural ordinary differential equations.
\newblock {\em Advances in neural information processing systems}, 31, 2018.

\bibitem{Yifan2021}
Yifan Chen, Bamdad Hosseini, Houman Owhadi, and Andrew~M Stuart.
\newblock Solving and learning nonlinear {PDE}s with {G}aussian processes.
\newblock {\em Journal of Computational Physics}, 447:110668, 2021.

\bibitem{cicci2023}
Ludovica Cicci, Stefania Fresca, Mengwu Guo, Andrea Manzoni, and Paolo Zunino.
\newblock Uncertainty quantification for nonlinear solid mechanics using reduced order models with {G}aussian process regression.
\newblock {\em Computers \& Mathematics with Applications}, 149:1--23, 2023.

\bibitem{costabal2019multi}
Francisco~Sahli Costabal, Paris Perdikaris, Ellen Kuhl, and Daniel~E Hurtado.
\newblock Multi-fidelity classification using {G}aussian processes: accelerating the prediction of large-scale computational models.
\newblock {\em Computer Methods in Applied Mechanics and Engineering}, 357:112602, 2019.

\bibitem{Damianou2013}
Andreas Damianou and Neil~D Lawrence.
\newblock Deep {G}aussian processes.
\newblock In {\em Proceedings of the 16th International Conference on Artificial Intelligence and Statistics}, volume~31 of {\em Proceedings of Machine Learning Research}, pages 207--215, 2013.

\bibitem{Fasel2022}
Urban Fasel, J~Nathan Kutz, Bingni~W Brunton, and Steven~L Brunton.
\newblock Ensemble-{SIND}y: {R}obust sparse model discovery in the low-data, high-noise limit, with active learning and control.
\newblock {\em Proceedings of the Royal Society A}, 478(2260):20210904, 2022.

\bibitem{ghattas2021learning}
Omar Ghattas and Karen Willcox.
\newblock Learning physics-based models from data: perspectives from inverse problems and model reduction.
\newblock {\em Acta Numerica}, 30:445--554, 2021.

\bibitem{Mark2021}
Mark Girolami, Eky Febrianto, Ge~Yin, and Fehmi Cirak.
\newblock The statistical finite element method (stat{FEM}) for coherent synthesis of observation data and model predictions.
\newblock {\em Computer Methods in Applied Mechanics and Engineering}, 375:113533, 2021.

\bibitem{gorry1990general}
Peter~A Gorry.
\newblock General least-squares smoothing and differentiation by the convolution (savitzky-golay) method.
\newblock {\em Analytical Chemistry}, 62(6):570--573, 1990.

\bibitem{guo2022bayesian}
Mengwu Guo, Shane~A McQuarrie, and Karen~E Willcox.
\newblock {B}ayesian operator inference for data-driven reduced-order modeling.
\newblock {\em Computer Methods in Applied Mechanics and Engineering}, 402:115336, 2022.

\bibitem{hansen2023learning}
Derek Hansen, Danielle~C Maddix, Shima Alizadeh, Gaurav Gupta, and Michael~W Mahoney.
\newblock Learning physical models that can respect conservation laws.
\newblock {\em arXiv preprint arXiv:2302.11002}, 2023.

\bibitem{Hirsh2022}
Seth~M Hirsh, David~A Barajas-Solano, and J~Nathan Kutz.
\newblock Sparsifying priors for {B}ayesian uncertainty quantification in model discovery.
\newblock {\em Royal Society Open Science}, 9(2):211823, 2022.

\bibitem{keith2021learning}
Brendan Keith, Akshay Khadse, and Scott~E Field.
\newblock Learning orbital dynamics of binary black hole systems from gravitational wave measurements.
\newblock {\em Physical Review Research}, 3(4):043101, 2021.

\bibitem{Liu2014}
Bo~Liu, Qingfu Zhang, and Georges~GE Gielen.
\newblock A {G}aussian process surrogate model assisted evolutionary algorithm for medium scale expensive optimization problems.
\newblock {\em IEEE Transactions on Evolutionary Computation}, 18(2):180--192, 2013.

\bibitem{MARREL2009742}
Amandine Marrel, Bertrand Iooss, Béatrice Laurent, and Olivier Roustant.
\newblock Calculations of {S}obol indices for the {G}aussian process metamodel.
\newblock {\em Reliability Engineering \& System Safety}, 94(3):742--751, 2009.

\bibitem{mcquarrie2021data}
Shane~A McQuarrie, Cheng Huang, and Karen~E Willcox.
\newblock Data-driven reduced-order models via regularised operator inference for a single-injector combustion process.
\newblock {\em Journal of the Royal Society of New Zealand}, 51(2):194--211, 2021.

\bibitem{meng2023sparse}
Rui Meng and Xianjin Yang.
\newblock Sparse {G}aussian processes for solving nonlinear {PDE}s.
\newblock {\em Journal of Computational Physics}, 490:112340, 2023.

\bibitem{murphy2012machine}
Kevin~P Murphy.
\newblock {\em Machine Learning: A Probabilistic Perspective}.
\newblock MIT Press, 2012.

\bibitem{Benjamin2016}
Benjamin Peherstorfer and Karen Willcox.
\newblock Data-driven operator inference for nonintrusive projection-based model reduction.
\newblock {\em Computer Methods in Applied Mechanics and Engineering}, 306:196--215, 2016.

\bibitem{pfortner2022physics}
Marvin Pf{\"o}rtner, Ingo Steinwart, Philipp Hennig, and Jonathan Wenger.
\newblock Physics-informed {G}aussian process regression generalizes linear {PDE} solvers.
\newblock {\em arXiv preprint arXiv:2212.12474}, 2022.

\bibitem{Qian2022}
Elizabeth Qian, Ionu\c{t}-Gabriel Farca\c{s}, and Karen Willcox.
\newblock Reduced operator inference for nonlinear partial differential equations.
\newblock {\em SIAM Journal on Scientific Computing}, 44(4):A1934--A1959, 2022.

\bibitem{RAISSI2017683}
Maziar Raissi, Paris Perdikaris, and George~Em Karniadakis.
\newblock Machine learning of linear differential equations using {G}aussian processes.
\newblock {\em Journal of Computational Physics}, 348:683--693, 2017.

\bibitem{williams2006gaussian}
Carl~E Rasmussen and Christopher~KI Williams.
\newblock {\em {G}aussian Processes for Machine Learning}.
\newblock The MIT Press, 2006.

\bibitem{Simo2011}
Simo S{\"a}rkk{\"a}.
\newblock Linear operators and stochastic partial differential equations in {G}aussian process regression.
\newblock In {\em Artificial Neural Networks and Machine Learning -- ICANN 2011: 21st International Conference on Artificial Neural Networks, Proceedings Part II 21}, pages 151--158. Springer, 2011.

\bibitem{shampine1997matlab}
Lawrence~F Shampine and Mark~W Reichelt.
\newblock The matlab ode suite.
\newblock {\em SIAM journal on scientific computing}, 18(1):1--22, 1997.

\bibitem{Naoya2017}
Naoya Takeishi, Yoshinobu Kawahara, Yasuo Tabei, and Takehisa Yairi.
\newblock Bayesian dynamic mode decomposition.
\newblock In {\em 26th International Joint Conference on Artificial Intelligence, IJCAI 2017}, pages 2814--2821, 2017.

\bibitem{thaler2019sparse}
Stephan Thaler, Ludger Paehler, and Nikolaus~A Adams.
\newblock Sparse identification of truncation errors.
\newblock {\em Journal of Computational Physics}, 397:108851, 2019.

\bibitem{uy2023operator}
Wayne Isaac~Tan Uy, Dirk Hartmann, and Benjamin Peherstorfer.
\newblock Operator inference with roll outs for learning reduced models from scarce and low-quality data.
\newblock {\em Computers \& Mathematics with Applications}, 145:224--239, 2023.

\bibitem{willcox2021imperative}
Karen~E Willcox, Omar Ghattas, and Patrick Heimbach.
\newblock The imperative of physics-based modeling and inverse theory in computational science.
\newblock {\em Nature Computational Science}, 1(3):166--168, 2021.

\bibitem{yang2021inference}
Shihao Yang, Samuel~WK Wong, and SC~Kou.
\newblock Inference of dynamic systems from noisy and sparse data via manifold-constrained {G}aussian processes.
\newblock {\em Proceedings of the National Academy of Sciences}, 118(15):e2020397118, 2021.

\bibitem{Ye2022}
Dongwei Ye, Pavel Zun, Valeria Krzhizhanovskaya, and Alfons~G. Hoekstra.
\newblock Uncertainty quantification of a three-dimensional in-stent restenosis model with surrogate modelling.
\newblock {\em Journal of The Royal Society Interface}, 19(187):20210864, 2022.

\bibitem{zhang2018robust}
Sheng Zhang and Guang Lin.
\newblock Robust data-driven discovery of governing physical laws with error bars.
\newblock {\em Proceedings of the Royal Society A: Mathematical, Physical and Engineering Sciences}, 474(2217):20180305, 2018.

\bibitem{zhuang2021model}
Qinyu Zhuang, Juan~Manuel Lorenzi, Hans-Joachim Bungartz, and Dirk Hartmann.
\newblock Model order reduction based on {R}unge-{K}utta neural networks.
\newblock {\em Data-Centric Engineering}, 2:e13, 2021.

\bibitem{Alvarez2009}
Mauricio Álvarez, David Luengo, and Neil~D. Lawrence.
\newblock Latent force models.
\newblock In {\em Proceedings of the 12th International Conference on Artificial Intelligence and Statistics}, volume~5 of {\em Proceedings of Machine Learning Research}, pages 9--16, 2009.

\end{thebibliography}

\end{document}